\documentclass{article}

\PassOptionsToPackage{numbers, compress}{natbib}

\usepackage[final]{neurips_2022}

\usepackage[utf8]{inputenc} 
\usepackage[T1]{fontenc}    
\usepackage{hyperref}       
\usepackage{url}            
\usepackage{booktabs}       
\usepackage{amsfonts}       
\usepackage{nicefrac}       
\usepackage{microtype}      
\usepackage{xcolor}         
\usepackage{multirow}
\usepackage{subfigure}
\usepackage{wrapfig}
\usepackage{enumerate}
\usepackage{graphicx}
\usepackage{amsmath}
\usepackage{amssymb,amsmath,amsthm,enumerate,threeparttable,bm,graphicx,subfigure}
\usepackage{bbm}
\usepackage{algorithm, algorithmic}
\usepackage{setspace}
\usepackage{makecell}

\title{Hierarchical Graph Transformer with Adaptive \\Node Sampling}

\author{%
	Zaixi Zhang$^{1,2}$ Qi Liu$^{1,2}$\thanks{Qi Liu is the corresponding author.}, Qingyong Hu$^3$, Chee-Kong Lee$^4$\\
	1: Anhui Province Key Lab of Big Data Analysis and Application,\\ School of Computer Science and Technology, University of Science and Technology of China\\2:State Key Laboratory of Cognitive Intelligence, Hefei, Anhui, China\\3:Hong Kong University of Science and Technology, 4: Tencent America\\
	zaixi@mail.ustc.edu.cn, qiliuql@ustc.edu.cn\\qhuag@cse.ust.hk,~ cheekonglee@tencent.com
}

\begin{document}

\maketitle

\begin{abstract}
The Transformer architecture has achieved remarkable success in a number of domains including natural language processing and computer vision. 
However, when it comes to graph-structured data, transformers have not achieved competitive performance, especially on large graphs. In this paper, we identify the main deficiencies of current graph transformers: 
(1) Existing node sampling strategies in Graph Transformers are agnostic to the graph characteristics and the training process. (2) Most sampling strategies only focus on local neighbors and neglect the long-range dependencies in the graph.
We conduct experimental investigations on synthetic datasets to show that existing sampling strategies are sub-optimal. To tackle the aforementioned problems, we formulate the optimization strategies of node sampling in Graph Transformer as an adversary bandit problem, where the rewards are related to the attention weights and can vary in the training procedure. Meanwhile, we propose a hierarchical attention scheme with graph coarsening to capture the long-range interactions while reducing computational complexity. Finally, we conduct extensive experiments on real-world datasets to demonstrate the superiority of our method over existing graph transformers and popular GNNs. Our code is 
open-sourced at https://github.com/zaixizhang/ANS-GT.
\end{abstract}
\section{Introduction}
In recent years, the Transformer architecture \cite{vaswani2017attention} and its variants (e.g., Bert \cite{devlin2018bert} and ViT \cite{dosovitskiy2020image}) have achieved unprecedented successes in natural language processing (NLP) and computer vision (CV). In light of the superior performance of Transformer, some recent works \cite{kreuzer2021rethinking, ying2021transformers} attempt to generalize Transformer for graph data by treating each node as a token and designing dedicated positional encoding. However, most of these works only focus on small graphs such as molecular graphs with tens of atoms \cite{ying2021transformers}. For instance, Graphormer \cite{ying2021transformers} achieves state-of-the-art performance on molecular property prediction tasks. When it comes to large graphs, the quadratic computational and storage complexity of the vanilla Transformer with the number of nodes inhibits the practical application. Although some Sparse Transformer methods \cite{roy2021efficient, beltagy2020longformer, kitaev2020reformer} can improve the efficiency of the vanilla Transformer, they have not
exploited the unique characteristics of graph data and require a quadratic or at least sub-quadratic
space complexity, which is still unaffordable in most practical cases. Moreover, the full-attention mechanism
potentially introduces noise from numerous irrelevant nodes in the full graph.

To generalize Transformer to large graphs, existing Transformer-based methods \cite{dwivedi2020generalization,zhao2021gophormer,devlin2018bert} on graphs explicitly or implicitly restrict each node’s receptive field to
reduce the computational and storage complexity. For example, Graph-Bert \cite{zhang2020graph} restricts the receptive field of each node to the nodes with top-$k$ intimacy scores such as Personalized PageRank (PPR). GT-Sparse \cite{dwivedi2020generalization} only considers 1-hop neighboring nodes. We argue that existing Graph Transformers have the following deficiencies: (1) The fixed node sampling strategies in existing Graph Transformers are ignorant of the graph properties, which may sample uninformative nodes for attention. Therefore, an adaptive node sampling strategy aware of the graph properties is needed. We conduct case studies in Section \ref{motivation} to support our arguments. (2) Though the sampling method enables scalability, most node sampling strategies focus on local neighbors and neglect the long-range dependencies and global contexts of graphs. Hence, incorporating complementary global information is necessary for Graph Transformer. 

To solve the challenge (1), we propose \textbf{A}daptive \textbf{N}ode \textbf{S}ampling for \textbf{G}raph \textbf{T}ransformer (ANS-GT) and formulate the optimization strategy of node sampling in Graph Transformer as an adversary bandit problem. 
Specifically in ANS-GT, we modify Exp4.P method \cite{beygelzimer2011contextual} to adaptively assign weights to several chosen sampling heuristics (e.g., 1-hop neighbors, 2-hop neighbors, PPR) and combine these sampling strategies to sample informative nodes.
The reward is proportional to the attention weights and the sampling probabilities of nodes, i.e. the reward to a certain sampling heuristic is higher if the sampling probability distribution and the node attention weights distribution are more similar. Then in the training process of Graph Transformer, the node sampling strategy is updated simultaneously to sample more informative nodes. With more informative nodes input into the self-attention module, ANS-GT can achieve better performance.

To tackle the challenge (2), we propose a hierarchical attention scheme for Graph Transformer to encode both local and global information for each node. The hierarchical attention scheme consists of fine-grained local attention and coarse-grained global attention. In the local attention, we use the aforementioned adaptive node sampling strategy to select informative local nodes for attention. As for global attention, we first use graph coarsening algorithms \cite{loukas2019graph} to pre-process the input graph and generate a coarse graph. Such algorithms mimic a down-sampling of the original graph via grouping the nodes into super-nodes while preserving global graph information as much as possible. The center nodes then interact with the sampled super-nodes. Such coarse-grained global attention helps each node capture long-distance dependencies while reducing the computational complexity of the vanilla Graph Transformers.   

We conduct extensive experiments on real-world datasets to show the effectiveness of ANS-GT. Our method outperforms all the existing Graph Transformer architectures and obtains state-of-the-art results on 6 benchmark datasets. Detailed analysis and ablation studies further show the superiority of the adaptive node sampling module and the hierarchical attention scheme. 

In summary, we  make the following contributions:
\begin{itemize}
    \item We propose \textbf{A}daptive \textbf{N}ode \textbf{S}ampling for \textbf{G}raph \textbf{T}ransformer (ANS-GT), which modifies a multi-armed bandit algorithm to adaptively sample nodes for attention.
    \item In the hierarchical attention scheme, we introduce coarse-grained global attention with graph coarsening, which helps graph transformer capture long-range dependencies while increasing efficiency. 
    \item We empirically evaluate our method on six benchmark datasets to show the advantage over existing Graph Transformers and popular GNNs.
\end{itemize}
\section{Related Work}
\subsection{Transformers for Graph}
Recently, Transformer \cite{vaswani2017attention} has shown its superiority in an increasing number of domains \cite{devlin2018bert, dosovitskiy2020image, zerveas2021transformer}, e.g. Bert \cite{devlin2018bert} in NLP and ViT \cite{dosovitskiy2020image} in CV.
Existing works attempting to generalize Transformer to graph data mainly focus on two problems: (1) How to design dedicated positional encoding for the nodes; (2) How to alleviate the quadratic computational complexity of the vanilla Transformer and scale the Graph Transformer to large graphs.
As for the positional encoding, GT \cite{dwivedi2020generalization} firstly uses Laplacian eigenvectors to enhance node features. Graph-Bert \cite{zhang2020graph} studies employing Weisfeiler-Lehman code to encode structural information. Graphormer \cite{ying2021transformers} utilizes centrality encoding to enhance node features while incorporating edge information with spatial (SPD-indexed
attention bias) and edge encoding.
SAN \cite{kreuzer2021rethinking} further replaces the static Laplacian eigenvectors with learnable positional encodings and designs an attention mechanism that distinguishes local connectivity.
For the scalability issue, one immediate idea is to restrict the number of attending nodes. For example, GAT \cite{velivckovic2017graph} and GT-Sparse \cite{dwivedi2020generalization} only consider the 1-hop neighboring nodes; Gophormer \cite{zhao2021gophormer} uses GraphSAGE \cite{hamilton2017inductive} sampling to uniformly sample ego-graphs with pre-defined maximum depth; Graph-Bert \cite{zhang2020graph} restricts the receptive field of each node to the nodes with top-k intimacy scores (e.g., Katz and PPR). However, these fixed node sampling strategies sacrifice the advantage of the Transformer architecture. SAC \cite{li2020sac} tries to use an LSTM edge predictor to predict edges for self-attention operations. However, the fact that LSTM can hardly be parallelized reduces the computational efficiency of the Transformer. 

\subsection{Sparse Transformers}
In parallel, many efforts have been devoted to reducing the computational complexity of the Transformer in the field of NLP \cite{liu2022transformer} and CV \cite{tay2020efficient}. In the domain of NLP, Longformer \cite{beltagy2020longformer} applies block-wise or strode patterns while only fixing on fixed neighbors. Reformer \cite{kitaev2020reformer} replaces dot-product attention by using approximate attention computation based on locality-sensitive hashing. Routing Transformer \cite{roy2021efficient} employs online k-means clustering on the tokens. Linformer \cite{wang2020linformer} demonstrates that the self-attention mechanism can be approximated by a low-rank matrix and reduces the complexity from $\mathcal{O}(n^2)$ to $\mathcal{O}(n)$. As for vision transformers, Swin Transformer \cite{liu2021swin} proposes the shifted windowing scheme which brings greater efficiency by limiting self-attention computation to non-overlapping local windows while also allowing for cross-window connection. Focal Transformer \cite{yang2021focal} presents a new mechanism incorporating both fine-grained local and coarse-grained global attention to capture short- and long-range visual dependencies efficiently. However, these sparse transformers do not take the unique graph properties into consideration.
\subsection{Graph Neural Networks and Node Sampling}
Graph neural networks (GNNs) \cite{kipf2016semi, hamilton2017inductive, han2021adaptive, zhang2022protgnn, zhang2021motif, song2022towards, zhang2021graph, zhang2022model} follow a message-passing schema that iteratively updates the representation of a node by aggregating representations from neighboring nodes.
When generalizing to large graphs, Graph Neural Networks face a similar scalability issue. This is mainly due to the uncontrollable neighborhood expansion in the aggregation stage of GNN. Several node sampling algorithms have been proposed to limit the neighborhood expansion, which mainly falls into node-wise sampling methods and layer-wise sampling methods. In node-wise sampling, each node samples $k$ neighbors from its sampling distribution, then the total number of nodes in the
$l$-th layer becomes $\mathcal{O}(k^l)$. GraphSage \cite{hamilton2017inductive} is one of the most well-known node-wise sampling methods with the uniform sampling distribution. GCN-BS \cite{liu2020bandit} introduces a variance reduced sampler based on multi-armed bandits. To alleviate the exponential neighbor
expansion $\mathcal{O}(k^l)$ of the node-wise samplers, layer-wise samplers define the sampling distribution as a probability of
sampling nodes given a set of nodes in the upper layer \cite{chen2018fastgcn, huang2018adaptive, zou2019layer}. From another perspective, these sampling methods can also be categorized into fixed sampling strategies \cite{hamilton2017inductive, chen2018fastgcn, zou2019layer} and adaptive strategies \cite{liu2020bandit,yoon2021performance}. 
However, none of the above sampling methods in GNNs can be directly applied in Graph Transformer as Graph Transformer does not follow the message passing schema.

\section{Preliminaries}
\subsection{Problem Definition}
Let $G=(A, X)$ denote the unweighted graph where $A \in \mathbb{R}^{n\times n}$ represents the symmetric
adjacency matrix with $n$ nodes, and $X \in \mathbb{R}^{n\times p}$ is the attribute  matrix of $p$ attributes per
node. The element $A_{ij}$ in the adjacency matrix equals to 1 if there exists an edge between node $v_i$ and node $v_j$, otherwise $A_{ij}=0$. The label of node $v_i$ is $y_i$. 
In the node classification problem, the classifier has the knowledge of the labels of a subset of nodes $V_L$. The goal of semi-supervised node classification is to
infer the labels of nodes in $V \backslash V_L$ by learning a classification function.

\subsection{Transformer Architecture}
The Transformer architecture consists of a series of Transformer layers \cite{vaswani2017attention}.
Each Transformer layer has two parts: a multi-head self-attention (MHA) module and a position-wise feed-forward network (FFN). Let $\mathbf{H}=\left[\boldsymbol{h}_{1}, \cdots, \boldsymbol{h}_{m}\right]^{\top} \in \mathbb{R}^{m \times d}$ denote the input to the self-attention module where $d$ is the hidden dimension, $\boldsymbol{h}_{i} \in \mathbb{R}^{d \times 1}$ is the hidden representation at position $i$, and $m$ is the number of positions. The MHA module firstly projects the input $\mathbf{H}$ to query-, key-, value-spaces, denoted as $\mathbf{Q}, \mathbf{K}, \mathbf{V}$, using three matrices $\mathbf{W}_{Q} \in \mathbb{R}^{d \times d_{K}}, \mathbf{W}_{K} \in \mathbb{R}^{d \times d_{K}}$ and $\mathbf{W}_{V} \in \mathbb{R}^{d \times d_{V}}$:
\begin{equation}
    \mathbf{Q}=\mathbf{H} \mathbf{W}_{Q}, \quad \mathbf{K}=\mathbf{H} \mathbf{W}_{K}, \quad \mathbf{V}=\mathbf{H} \mathbf{W}_{V}.
\end{equation}
Then, in each head $i \in \{1,2,\dots, B\}$ ($B$ is the total number of heads), the scaled dot-product attention mechanism is applied to the corresponding $\{\mathbf{Q}_i, \mathbf{K}_i, \mathbf{V}_i\}$ :
\begin{equation}
   \text { head}_{i} = \operatorname{Softmax}\left( \frac{\mathbf{Q}_i \mathbf{K}_i^T}{\sqrt{d_{K}}}\right)  \mathbf{V}_i.
\end{equation}
Finally, the outputs from different heads are further concatenated and transformed to obtain the final output of MHA:
\begin{equation}
    \operatorname{MHA}(\mathbf{H})=\operatorname{Concat}\left(\text { head}_{1}, \ldots, \text {head}_{B}\right) \mathbf{W}_{O},
\end{equation}
where $\mathbf{W}_{O} \in \mathbb{R}^{d \times d}$. In this work, we employ $d_K=d_V=d/B$.

\subsection{Graph Coarsening}
The goal of Graph Coarsening \cite{ron2011relaxation, loukas2019graph, huang2021scaling} is to reduce the number of nodes in a graph by clustering them into super-nodes while preserving the global information of the graph as much as possible. Given a graph $G=(V,E)$ ($V$ is the node set and $E$ is the edge set), the coarse graph is a smaller weighted graph $G' = (V', E')$. $G'$ is obtained from the original graph by first computing a partition $\{C_1, C_2, \cdots, C_{|V'|}\}$ of $V$, i.e., the clusters $C_1\cdots C_{|V'|}$ are disjoint and cover all the nodes in $V$. Each cluster $C_i$ corresponds to a super-node in $G'$ 
The partition can also be characterized by a matrix $\hat{P}\in\{0,1\}^{|V|\times |V'|}$, with $\hat{P}_{ij} = 1$ if and only if node $v_i$ in $G$ belongs to cluster $C_j$. Its normalized version can be defined by $P \triangleq \hat{P}D^{-\frac{1}{2}}$, where $D$ is a $|V'| \times |V'|$ diagonal matrix with $|C_i|$ as its $i$-th diagonal entry. The feature matrix and weighted adjacency matrix of $G'$ are defined by $X'\triangleq P^TX$ and $A'\triangleq P^T AP$.
After Graph Coarsening, the number of nodes/edges in $G'$ is significantly smaller than that of $G$. The coarsening rate can be defined as $c = \frac{|V'|}{|V|}$.
\section{Motivating Observations}\label{motivation}
To generalize Transformer to large graphs, existing Graph Transformer models typically choose to sample a batch of nodes for attention. However, real-world graph datasets exhibit different properties, which makes a fixed node sampling strategy unsuitable for all kinds of graphs. Here, we present a simple yet intuitive case study to illustrate how the performance of Graph Transformer changes with different node sampling strategies. The main idea is to use four popular node sampling strategies for node classification: 1-hop neighbors, 2-hop neighbors, PPR, and KNN. Then, we will check the performance of Graph Transformer on graphs with different properties. If the performance drops sharply when the property varies, it will indicate that graphs with different properties may require different node sampling strategies.

\begin{wrapfigure}{r}{6cm}
\centering
\vspace{-1em}
	\includegraphics[width=\linewidth]{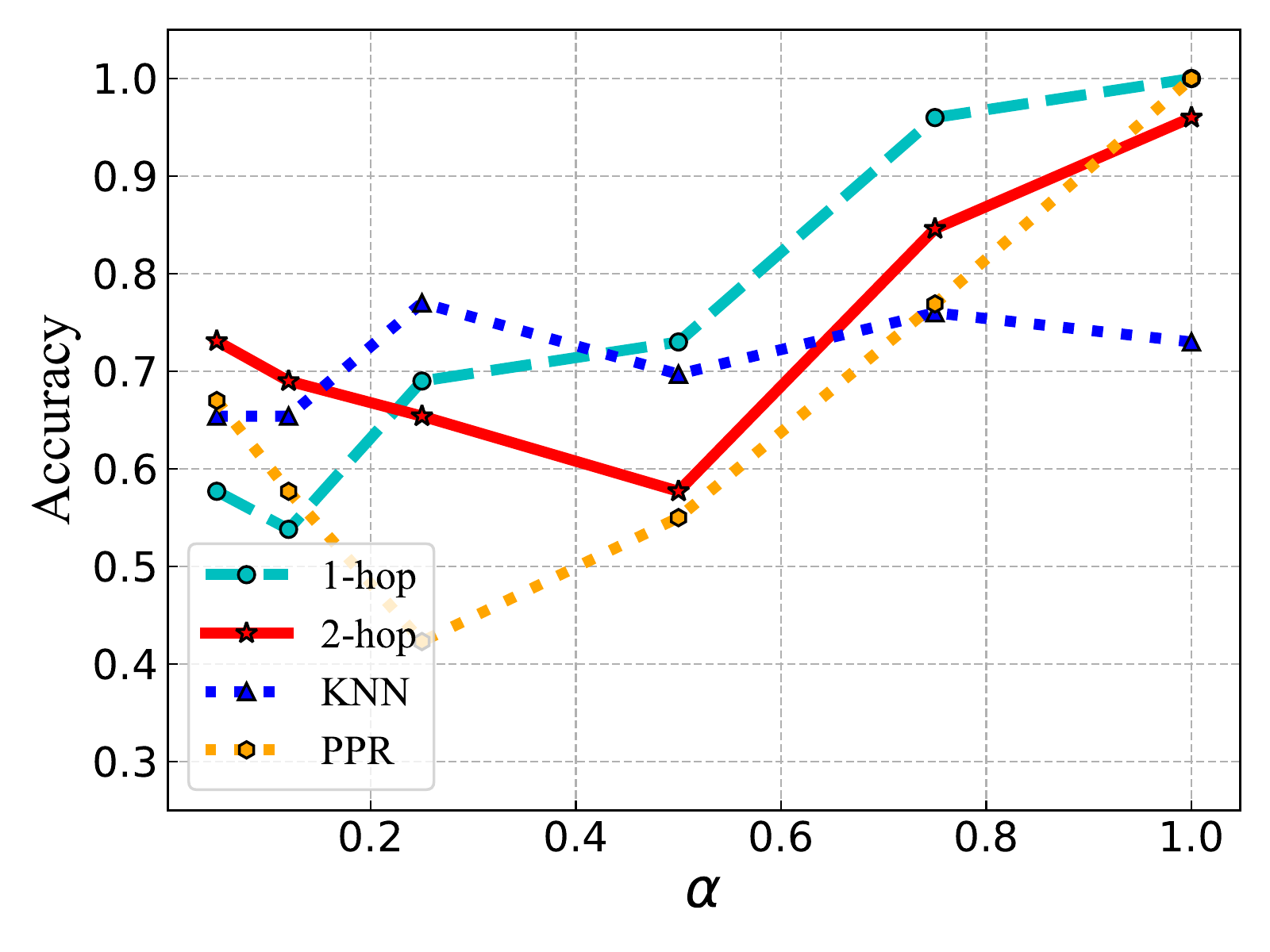}
	\vspace{-2.0em}
    \caption{Performance of Graph Transformer using different node sampling mechanisms: 1-hop, 2-hop, PPR and kNN respectively on Newman networks.}
    \vspace{-2em}
    \label{motivation1}
\end{wrapfigure}

We conduct experiments on the Newman artificial networks \cite{girvan2002community} since it enable us to obtain networks with different properties easily. More detailed settings can be found in Appendix \ref{details of observation}. Here, we consider the property of homophily/heterophily as one example. Following previous works \cite{zhu2020beyond}, the degree of homophily $\alpha$ can be defined as the fraction of edges in a network connecting nodes with the same class label, i.e. $\alpha \triangleq \frac{|(v_i, v_j) \in E \wedge y_i = y_j|}{|E|}$.
Graphs with $\alpha$ closer to 1 tend to have more edges connecting
nodes within the same class, i.e. strong homophily; whereas networks with $\alpha$ closer to 0 have more
edges connecting nodes in different classes, i.e. strong heterophily.

As shown in Figure \ref{motivation1}, there is no sampling strategy that dominates other strategies across the spectrum of $\alpha$, which supports our claim that an adaptive sampling method is needed for graphs with different properties. For graphs with strong homophily (e.g., $\alpha = 1.0$), it is easy to obtain high accuracy by sampling 1-hop neighbors or nodes with top PPR scores. On the other hand, for graphs with strong heterophily (e.g., $\alpha = 0.05$), the accuracy of using 2-hop neighbors as neighborhoods (i.e., $73.1\%$) is
much higher than that of using 1-hop neighbors and PPR. The probable reason is that the homophily ratio of 2-hop neighbors may rise with the increase of inter-class edges. Finally, Graph Transformer with KNN node sampling can get the most consistent results since KNN only calculates the similarity of attributes. Thus, KNN achieves the best performance (i.e., $77.2\%$ accuracy) when all the nodes are connected randomly, i.e. $\alpha = 0.25$ for the Newman network.

Hence, considering different graph properties (e.g., homophily/heterophily), Graph Transformer should adaptively sample the most informative nodes for attention. These observations motivate us to design the proposed hierarchical Graph Transformer with adaptive node sampling in Section \ref{method}.

\section{The Proposed Method}\label{method}
\begin{figure*}[t]
    \centering
    \includegraphics[width=0.99\linewidth]{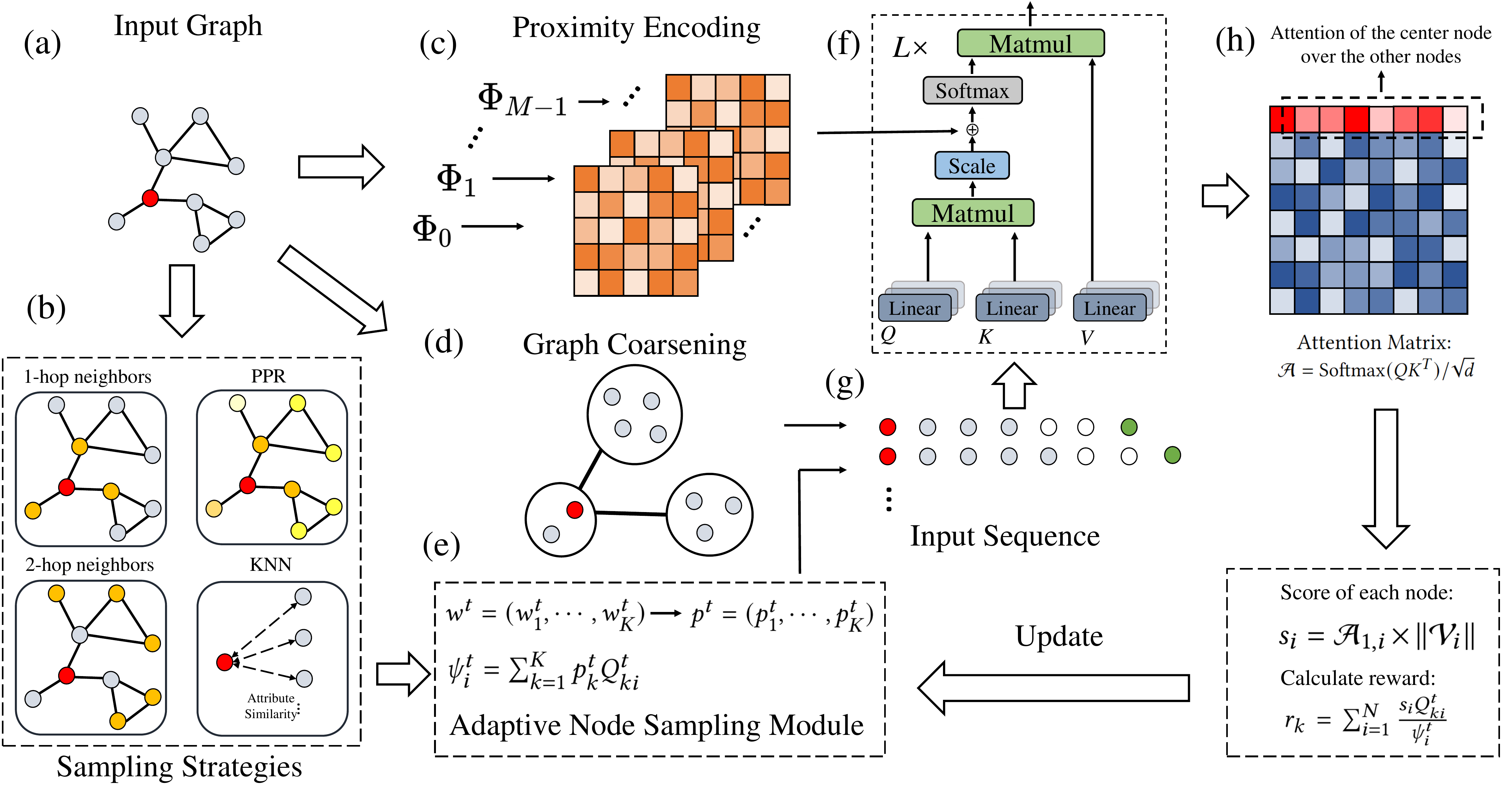}
    \caption{Model framework of our proposed method: (a) An example input graph. The center node for sampling is colored red. (b) We consider four sampling strategies in this work, i.e. 1-hop neighbors, 2-hop neighbors, PPR, and KNN. (c) The proximity encoding module. (d) Graph coarsening to cluster nodes into super-nodes. (e) The adaptive node sampling module. (f) The self-attention module in Graph Transformer. The output node embeddings are used for node classification. (g) In the sampled input node sequences, the gray nodes are the fine-grained nodes; the white nodes are the coarse-grained nodes from graph coarsening; the green nodes denote the global nodes. (h) We use the first row of the attention matrix, i.e., $\mathcal{A}_{1,i}$ multiplying the magnitude of the corresponding value $V_i$ to represent the significance of each node. Then we calculate the reward for each sampling strategy and update the weights.}
    \label{overview}
\end{figure*}

In light of the limitations of existing Graph Transformers for large graphs and the motivating observation in the previous section, we propose two effective methods for Graph Transformer to adaptively sample informative nodes and capture the long-range coarse-grained dependencies in this section. We also show that our method has a computational complexity of $\mathcal{O}(n)$. The overview of the model framework is shown in Figure \ref{overview}.
\subsection{Adaptive Node Sampling}
Our Adaptive Node Sampling module aims to adaptively choose the batch of most informative nodes by a multi-armed bandit mechanism. In our setting, it is intuitive that the contributions of nodes to the learning performance can be time-sensitive. Meanwhile, the rewards which are associated with the model training process are not independent random variables across the iterations. The above situation can satisfy the adversarial setting in the multi-armed bandit problem \cite{auer2002nonstochastic}. To adaptively choose the most informative nodes with the
designed sampling strategies, we adjust the method ALBL proposed in \cite{hsu2015active}, which modifies the EXP4.P
method \cite{beygelzimer2011contextual}. EXP4.P possesses a strong theoretical guarantee for the adversarial setting.

Formally, let $w^t = (w_1^t, \cdots, w_K^t)$ be the adaptive weight vector in iteration $t$, where the $k$-th non-negative element $w_k^t$ is the weight corresponding to the $k$-th node sampling strategy. The weight vector $w^t$ is then scaled to a probability vector $p^t = (p_1^t, \cdots, p_K^t)$ where $p_k^t \in [p_{min}, 1]$ with $p_{min} > 0$. Our ANS-GT adaptively sample nodes based on the probability vector and then obtains the reward of the action. 

For each center node, we consider the sampling probability matrix $Q^t \in \mathbb{R}^{K \times n}$, where $K$ is the number of sampling heuristics and $n$ is the number of nodes in the graph. Specifically, $Q_{k,j}^t$ denotes the $k$-th sampling strategy's preference on selecting node $j$ in iteration $t$ and $Q^t$ is normalized to satisfy $\sum_{j=1}^n Q_{k,j}^t = 1$. Note that our ANS-GT is a general framework and is not restricted to a certain set of sampling heuristics. In our work, we adopt four representative sampling heuristics:

\textbf{1-/2-hop neighbors}: We adopt the normalized adjacency matrix $\widetilde{A} = \hat D^{-\frac{1}{2}} \hat A \hat D^{-\frac{1}{2}}$ for 1-hop neighbors and $\widetilde{A}^2$ for 2-hop neighbors, where $\hat A = A + I$ is the adjacency matrix of the graph $G$ with self connections added and $\hat D$ is a diagonal matrix with $\hat D_{ii} = \sum_j \hat A_{ij}$.

\textbf{KNN}: We adopts the cosine similarity of node attributes to measure the similarities of nodes. Mathematically, the similarity score $S_{ij}$ between node $i$ and $j$ is calculated as $S_{ij} = x_i \cdot x_j/ (|x_i|\cdot|x_j|)$ where $x_i$ is the feature vector of node $v_i$.

\textbf{PPR}: The Personalized PageRank \cite{page1999pagerank} matrix $S$ is calculated as: $S = c (I -(1-c)\overline{A})$, where factor $c \in [0, 1]$  (set to 0.15 in our experiments). $\overline{A}$ denotes the column-normalized adjacency matrix.

Finally, given the probability vector $p^t$ and the node sampling matrices $Q^t$, the final node sampling probability is:
\begin{equation}
    \psi^t_i = \sum_{k=1}^K p_k^t Q^t_{ki}.
    \label{probability}
\end{equation}

We introduce a novel reward scheme based on the attention weights which is intrinsic in Transformer. Formally, given an attention matrix $\mathcal{A} = \operatorname{Softmax}(QK^T)/\sqrt{d}$, we use the first row of the attention matrix, i.e., $\mathcal{A}_{1,i}$ multiplying the magnitude of corresponding value $V_i$ to represent the significance of each node to the center node: $s_i = \mathcal{A}_{1,i} \times \|\mathcal{V}_i\|$. In the situation of multiple heads and layers in the Transformer, we average the significance scores in multiple attention matrices. The reward to the $k$-th sampling strategy is: $r_k = \sum_{i=1}^N \frac{s_i Q^t_{ki}}{\psi^t_i}$, where $N$ is the number of sampled nodes for each center node. $r_k$ can be interpreted as the dot product between the significance score vector and the normalized sampling probability vector. The intuition behind the reward design is that the reward to a certain sampling heuristic is higher if the sampling probability distribution and the node significance score distribution are closer. Thus, exploiting such a sampling heuristic can help graph transformer sample more informative nodes. Finally, we update  $w^t$ with the reward. In experiments, for the efficiency and stability of training, we update the sampling weights and resample nodes every $T$ epochs. The pseudo-code of ANS-GT is listed in Algorithm \ref{pseudo code}.

\subsection{Hierarchical Graph Attention}
We argue that most node sampling strategies (e.g., 1-hop neighbors) focus on local information and neglect long-range dependencies or global contexts. 
Therefore, to efficiently capture both the local and global information in the graph, we propose a novel hierarchical graph attention scheme including fine-grained attention and coarse-grained attention. Specifically, we use the proposed adaptive node sampling for local fined-grained attention. On the other hand, we adopt the graph coarsening algorithm \cite{loukas2019graph} to generate the coarsened graph $G'$. The sampled $n_s$ super-nodes from $G'$ are used to capture long-range dependencies. Similar to \cite{zhao2021gophormer}, we also use $n_g$ global nodes with learnable features to store global context information. Finally, the hierarchical nodes are concatenated with the center nodes as the input sequences. 

For the positional encoding, we use the proximity encoding in \cite{zhao2021gophormer}: $\Phi_m(v_i, v_j) = \widetilde{A}^m[i,j], m \in \{0, \cdots, M-1\}$, where $\widetilde{A}$ denotes the normalized adjacency matrices with
self-loop. 
Note that our framework is agnostic to the positional encoding scheme and we left other positional encoding methods such as Laplacian eigenvectors \cite{dwivedi2020generalization} for future exploration.
We follow the Graphormer framework to obtain the output of the $l$-th transformer layer, $\mathbf{H}^{(l)}$:
\begin{align}
    \hat {\mathbf{H}}^{(l-1)}&=\operatorname{MHA(LN}(\mathbf{H}^{(l-1)}))+\mathbf{H}^{(l-1)}\\
    {\mathbf{H}}^{(l)}&=\operatorname{FFN(LN}(\hat{\mathbf{H}}^{(l-1)}))+\hat{\mathbf{H}}^{(l-1)}.
\end{align}
We apply the layer normalization (LN) before the multi-head
self-attention (MHA) and the feed-forward network (FFN).

\begin{algorithm}[t]
\caption{ANS-GT}
\label{pseudo code}
\leftline{\textbf{Input}: Total training epochs $E$;  ~$p_{min}$; update period $T$; the number of sampled nodes $N$.}
\leftline{\textbf{Output}: Trained Graph Transformer model, optimized $w^t$.}
\begin{algorithmic}[1]
\STATE{Set $w^1_k = 1$ for $k= 1, \cdots, K$.} 
\STATE{Calculate the sampling probability matrix $Q^t$.} 
\FOR{$t = 1, 2, \cdots, E$}
\STATE Train Graph Transformer with the sampled node sequences.
\IF {$t\%T = 0$.}  
\STATE Obtains the attention matrices and calculate the significance scores: $s_i = \mathcal{A}_{1,i} \times \|\mathcal{V}_i\|$.
\STATE Set $W^t = \sum_{k=1}^K w_k^t$, and set $p_k^t = (1-K p_{min})\sum_{j=1}^K \frac{w_j^t}{W^t} + p_{min}$ for $k= 1, \cdots, K$.
\STATE Calculate $\psi^t_i$ in Equ. \ref{probability} and sample $N$ nodes.
\STATE Set $r_k = \sum_{i=1}^N \frac{s_i Q^t_{ki}}{\psi^t_i}$ and update the weight vector $w_k^{t+1}$ using \\$w_k^{t+1} = w_k^t e^{(\frac{p_{min}}{2})(r_k + \frac{1}{P_k^t})\sqrt{\frac{ln(N/0.1)}{KT}}}$.

\ENDIF
\ENDFOR
\end{algorithmic}
\end{algorithm}

\subsection{Optimization and Inference}
In the training and inference, we sample $\mathcal{S}$ input sequences for each center node and use the center node representation from the final Transformer layer $z_c^{(s)}$ for prediction. Note that the computational complexity is controllable by choosing suitable number of sampled nodes. A MLP (Multi-Layer Perceptron) is used to predict the node class:
\begin{equation}
  \widetilde{\boldsymbol{y}}^{(s)}=f_{MLP}\left(\boldsymbol{z}_c^{(s)} \right), 
\end{equation}
where $\widetilde{\boldsymbol{y}}_c \in \mathbb{R}^{C \times 1}$ stands for the classification result, $C$ stands for the number of classes. In the training process, we optimize the average cross entropy loss of labeled training nodes $V_L$:
\begin{equation}
    \mathcal{L}=-\frac{1}{\mathcal{S}}  \sum_{v_i\in V_L }\sum_{s=1}^{\mathcal{S}}\boldsymbol{y}_i^T \log \widetilde{\boldsymbol{y}}_i^{(s)},
\end{equation}
where $\boldsymbol{y}_i \in \mathbb{R}^{C \times 1}$ is the ground truth label of center node $v_i$. In the inference stage, we take a bagging aggregation to improve accuracy and reduce variance:
\begin{equation}
\widetilde{\boldsymbol{y}_i} =\frac{1}{\mathcal{S}} \sum_{s=1}^{\mathcal{S}} \widetilde{\boldsymbol{y}}_i^{(s)}.
\end{equation}
\subsection{Computational Complexity}
Compared with existing Graph Transformers, ANS-GT requires extra computational costs on graph coarsening and sampling weights update. Here we want to show that the overall computational complexity of ANS-GT is linear with the number of nodes $n$. Hence, ANS-GT is scalable to large graphs. First, the computational complexity of graph coarsening is linear with $n$ \cite{loukas2019graph} and we only need to do it once before training. Second, the computational cost of self-attention calculation in one epoch is $\mathcal{O}(n\mathcal{S}(N+ n_s + n_g)^2)$ where the number of sampled nodes $N$, the number of sampled super-nodes $n_s$, the number of global nodes $n_g$, and the number of augmentations $\mathcal{S}$ are specified constants. Finally, the cost of updating sampling weights is linear to $n$, which is mainly attributed to the computation of rewards. Empirical efficiency analyses of ANS-GT are shown in the Appendix \ref{additional analysis}.

\section{Experiments}
\subsection{Experimental Setup}
\begin{table}[t]
\centering
\scriptsize
\tabcolsep=0.15cm
\caption{Node classification performance (mean±std\%, the best results are bolded).}
\begin{tabular}{@{}c|ccccccccc@{}} \toprule
Model                              & Cora                & Citeseer            & Pubmed         & Chameleon              & Actor                & Squirrel & Texas	&Cornell&	Wisconsin             \\  \midrule
GCN                                & 87.33{\tiny±0.38}          & 79.43{\tiny±0.26}          & 84.86{\tiny±0.19}          & 60.96{\tiny±0.75}          & 31.39{\tiny±0.23}          & 43.15{\tiny±0.18} & 75.16{\tiny±0.95}& 66.74{\tiny±1.39}& 64.31{\tiny±2.16}         \\
GAT                                & 86.29{\tiny±0.53}          & 80.13{\tiny±0.62}          & 84.40{\tiny±0.05}          & 63.90{\tiny±0.46}          & 36.05{\tiny±0.35}          & 42.72{\tiny±0.39}  & 78.76{\tiny±0.87}& 76.04{\tiny±1.35}& 66.01{\tiny±3.48}        \\
GraphSAGE                          & 86.90{\tiny±0.84}          & 79.23{\tiny±0.53}          & 86.19{\tiny±0.18}          & 62.15{\tiny±0.42}          & 38.55{\tiny±0.46}          & 41.26{\tiny±0.26}  & 79.03{\tiny±1.44}& 72.54{\tiny±1.50}& 79.41{\tiny±3.60}        \\
APPNP              & 87.15{\tiny±0.43}          & 79.33{\tiny±0.35}          & 87.04{\tiny±0.17}          & 51.91{\tiny±0.56}          & 38.80{\tiny±0.25}          & 37.76{\tiny±0.45}    & 91.18{\tiny±0.75}& 91.75{\tiny±0.72}& 82.56{\tiny±3.57}        \\
JKNet                              & 87.70{\tiny±0.65}         & 78.43{\tiny±0.31}          & 87.64{\tiny±0.26}         & 62.92{\tiny±0.49}          & 33.42{\tiny±0.28}          & 42.60{\tiny±0.50}    & 77.51{\tiny±1.72}& 64.26{\tiny±1.16}& 81.20{\tiny±1.96}          \\
H$_2$GCN &  87.92{\tiny±0.82}         & 77.60{\tiny±0.76}          & 89.55{\tiny±0.14}         & 61.20{\tiny±0.95}          & 36.22{\tiny±0.33}          & 38.51{\tiny±0.20}  & 86.37{\tiny±2.67}& 84.93{\tiny±1.89}& 87.73{\tiny±1.57}        \\
GPRGNN & 88.27{\tiny±0.40}         & 78.46{\tiny±0.88}          & 89.38{\tiny±0.43}         & 64.56{\tiny±0.59}          & 39.27{\tiny±0.21}          & \bf46.34{\tiny±0.77}   &91.84{\tiny±1.25} &90.25{\tiny±1.93} &86.58{\tiny±2.58}       \\\midrule
GT	&71.84{\tiny±0.62}	    &67.38{\tiny±0.76}	    &82.11{\tiny±0.39}	    &57.86{\tiny±1.20}	    &37.94{\tiny±0.26}	    &25.68{\tiny±0.22} &66.70{\tiny±1.13} &60.39{\tiny±1.66} &65.08{\tiny±4.37}\\
SAN         & 74.02{\tiny±1.01}          & 70.64{\tiny±0.97}          & 86.22{\tiny±0.43}          & 55.62{\tiny±0.43}          & 38.24{\tiny±0.53}          & 25.56{\tiny±0.23}  &70.10{\tiny±1.82} &61.20{\tiny±1.17} &65.30{\tiny±3.80}        \\
Graphormer & 72.85{\tiny±0.76}          & 66.21{\tiny±0.83}          & 82.76{\tiny±0.24}          & 36.81{\tiny±1.96}          & 37.85{\tiny±0.29}          & 25.45{\tiny±0.12}    &68.56{\tiny±1.74} &59.41{\tiny±1.21} &67.53{\tiny±3.38}    \\ 
Gophormer                         & 87.65{\tiny±0.20} & 76.43{\tiny±0.78} & 88.33{\tiny±0.44} & 57.40{\tiny±0.14} & 37.50{\tiny±0.42} & 37.85{\tiny±0.36} &88.25{\tiny±1.96} &89.46{\tiny±1.51} &85.09{\tiny±2.60} \\
\midrule
ANS-GT                      & \textbf{88.60{\tiny±0.45}} & \textbf{80.25{\tiny±0.39}} & \textbf{89.56{\tiny±0.55}} & \bf 65.42{\tiny±0.71} & \textbf{40.10{\tiny±1.12}} & 45.88{\tiny±0.34}& \bf93.24{\tiny±1.85}&	\bf92.10{\tiny±1.78}&	\bf88.62{\tiny±2.24}\\ \bottomrule
\end{tabular}
\label{tab:result-1}
\end{table}

\textbf{Datasets. }To comprehensively evaluate the effectiveness of ANS-GT, we conduct experiments on the six benchmark datasets including citation graphs Cora, CiteSeer, and PubMed \cite{kipf2016semi}; Wikipedia graphs Chameleon, Squirrel; the Actor co-occurrence graph \cite{chien2020adaptive}; and WebKB datasets \cite{Pei2020Geom-GCN} including Cornell, Texas, and Wisconsin. We set the train-validation-test split as 60\%/20\%/20\%. The statistics of datasets are shown in the Appendix \ref{additional analysis}. 

\textbf{Baselines. }
To evaluate the effectiveness of ANS-GT on graph representation learning, we compare it with 12 baseline methods, including 8 popular GNN methods,
i.e. GCN~\cite{kipf2016semi}, GAT~\cite{velivckovic2017graph}, GraphSAGE~\cite{hamilton2017inductive}, JKNet~\cite{xu2018representation}, APPNP~\cite{klicpera2018predict}, Geom-GCN~\cite{Pei2020Geom-GCN}, H$_2$GCN~\cite{NEURIPS2020_58ae23d8}, and GPRGNN~\cite{DBLP:conf/iclr/ChienP0M21} along with four state-of-the-art Graph Transformers, i.e. GT~\cite{dwivedi2020generalization}, SAN~\cite{kreuzer2021rethinking}, Graphormer~\cite{ying2021transformers}, and Gophormer~\cite{zhao2021gophormer}. We use node classification accuracy as the evaluation metric.

\textbf{Implementation Details. }
We adopt AdamW as the optimizer and set the hyper-parameter $\epsilon$ to 1e-8 and ($\beta1,\beta2$) to (0.99,0.999). The peak learning rate is set to 2e-4 with a 100 epochs warm-up stage followed by a linear decay learning rate scheduler. We adopt the Variational Neighborhoods \cite{loukas2019graph} with a coarsening rate of 0.01 as the default coarsening method. All models were trained on one NVIDIA Tesla V100 GPU. 

\textbf{Parameter Settings.}
In the default setting, the dropout rate is set to 0.5, the end learning rate is set to 1e-9, the hidden dimension $d$ is set to 128, the number of training epochs is set to 1,000, update period $T$ is set to 100, $N$ is set to 20, $M$ is set to 10, and the number of attention head $H$ is set as 8. We tune other hyper-parameters on each dataset based on by grid search. The searching space of 
batch size, number of data augmentation $\mathcal{S}$, the number of layers $L$, number of sampled nodes, number of sampled super-nodes, number global nodes are $\{8, 16, 32\}$, $\{4, 8, 16, 32\}$, $\{2, 3, 4, 5, 6\}$, $\{10, 15, 20, 25\}$, $\{0, 3, 6, 9\}$, $\{1, 2, 3\}$ respectively. 

\begin{figure*}[t]
	\centering
	\subfigure[Cora]{\includegraphics[width=0.24\linewidth]{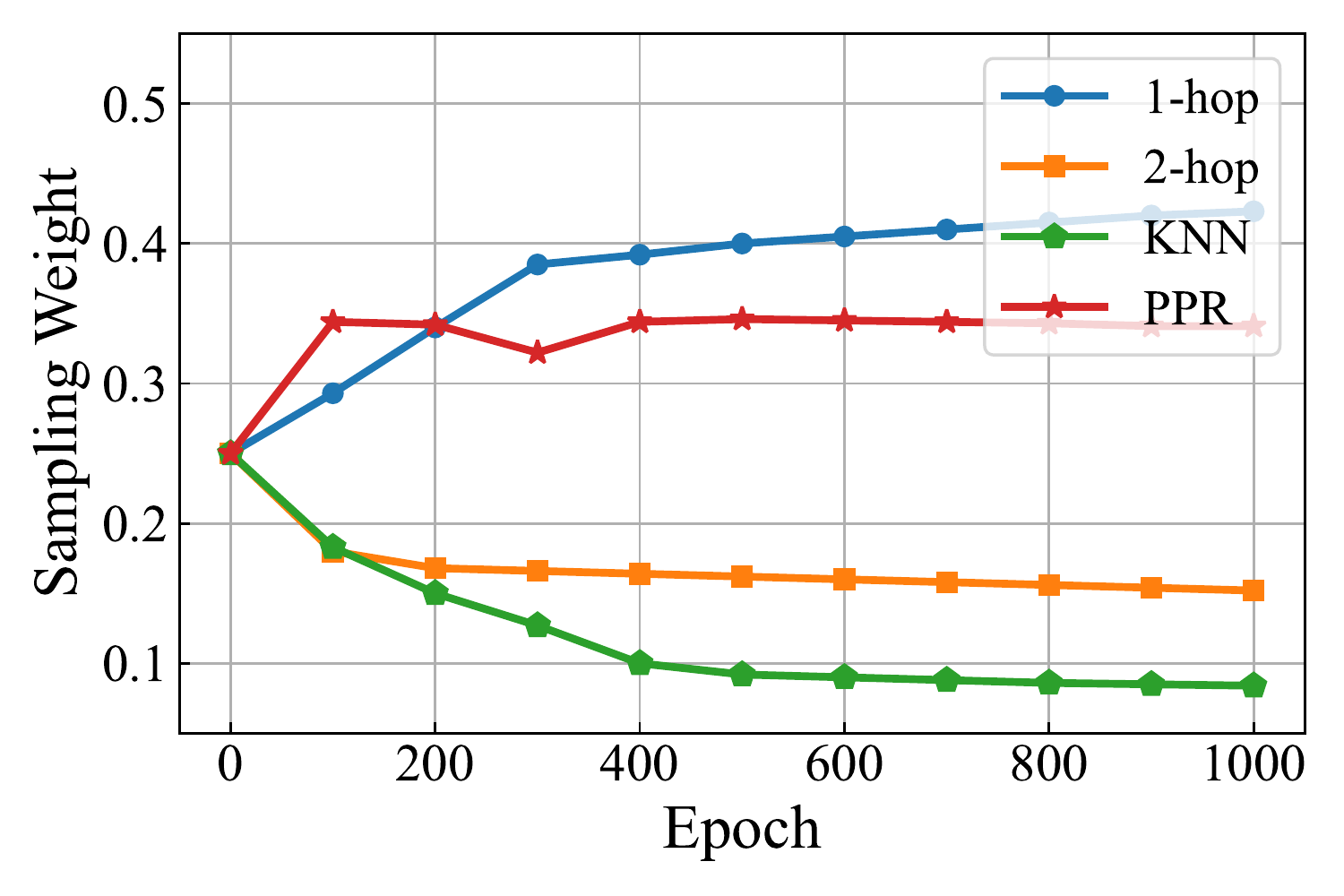}}
    \subfigure[Citeseer]{\includegraphics[width=0.24\linewidth]{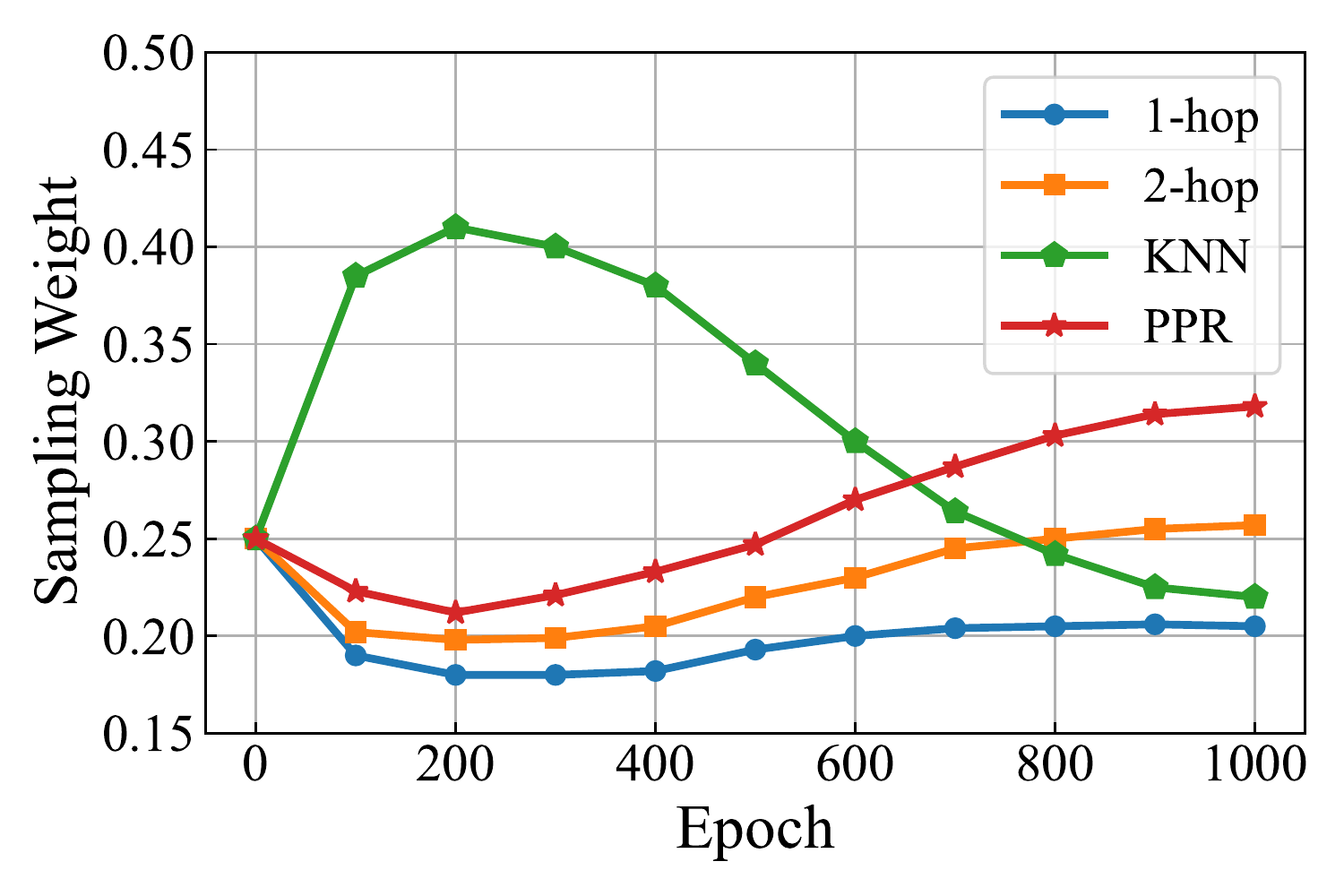}}
    \subfigure[Actor]{\includegraphics[width=0.24\linewidth]{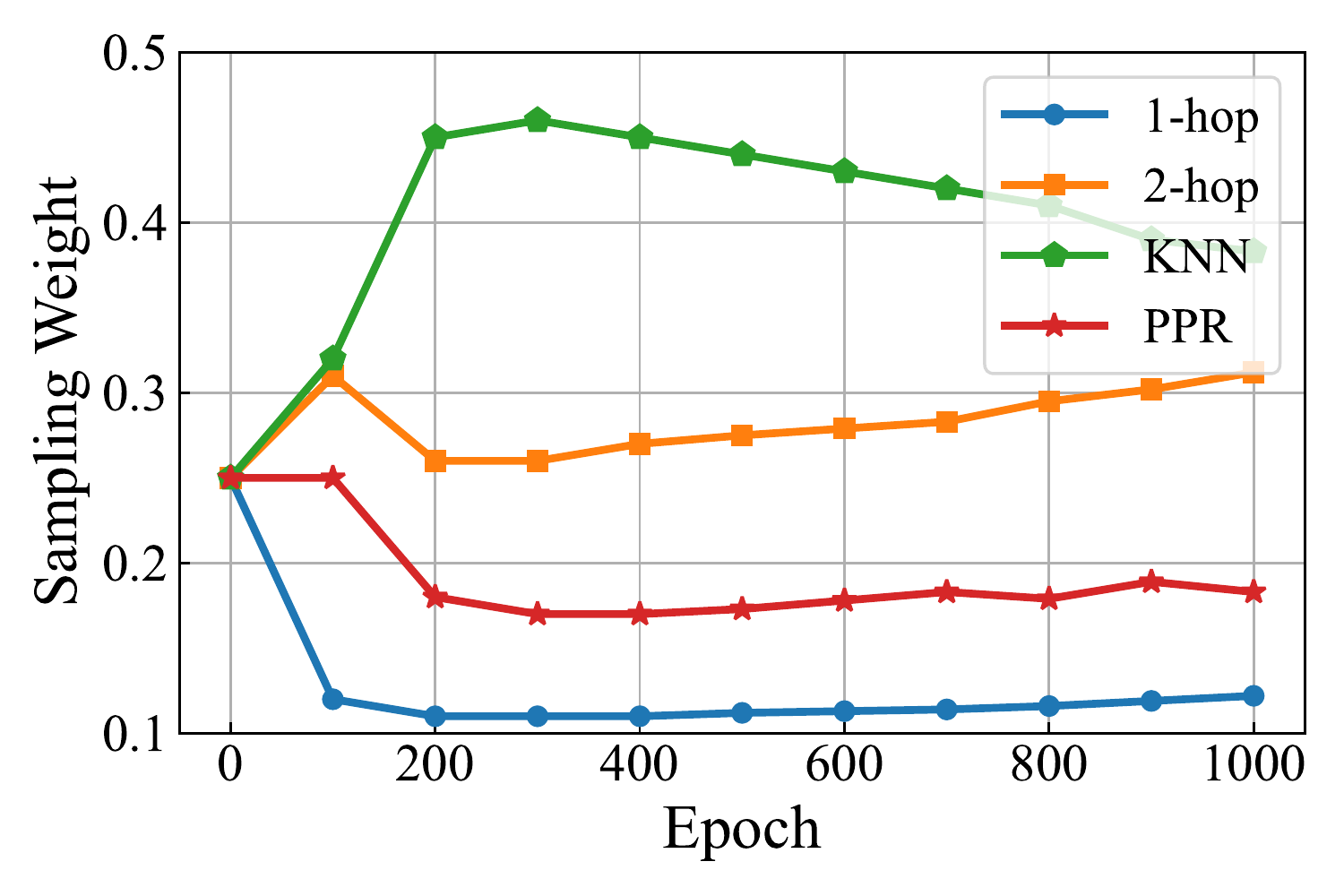}}
    \subfigure[Squirrel]{\includegraphics[width=0.24\linewidth]{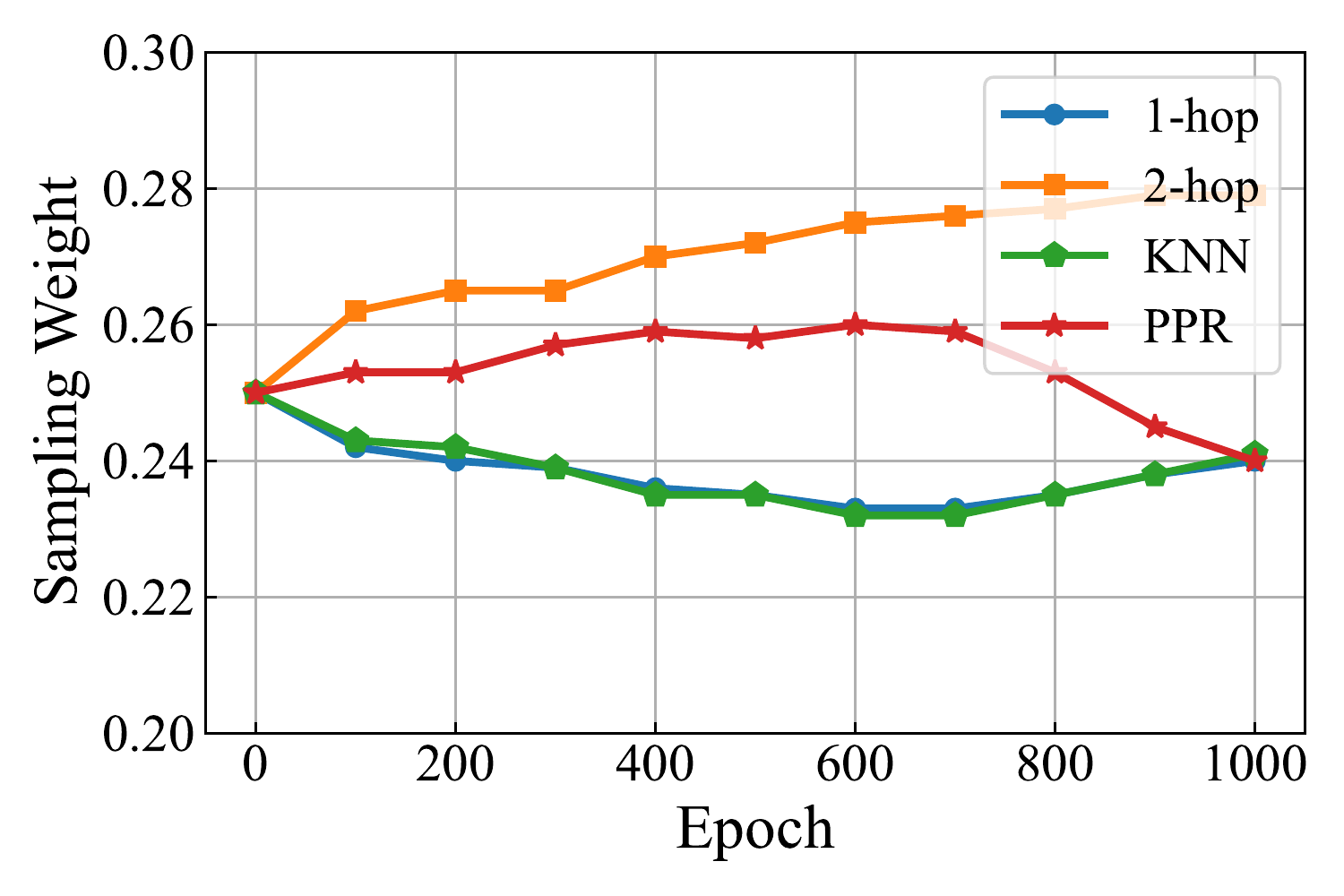}}
	\caption{The normalized sampling weights as a function of training epochs on four datasets.}
	\label{curve}
\end{figure*}
\subsection{Effectiveness of ANS-GT}
\textbf{Node Classification Performance. }
The node classification results are shown in Table \ref{tab:result-1}. We apply 3 independent runs on random data splitting and report the means and standard deviations. We have the following obervations: (1) Generally, we observe that ANS-GT overperforms all Graph Transformer baselines and achieves state-of-the-art results on all 6 datasets, which demonstrates the effectiveness of our proposed model. (2) We note that some Graph Transformer baselines achieve poor performance on node classification (e.g., GT only obtains $25.68\%$ on Squirrel) compared with graph neural network models. This is probably due to the full graph attention mechanisms or the fixed node sampling schemes of existing Graph Transformers. For instance, ANS-GT achieves an accuracy of 45.88$\%$ on Squirrel while the best baseline has 43.15$\%$.
\begin{wrapfigure}{r}{6cm}
\centering
\vspace{1.0em}
	\includegraphics[width=0.95\linewidth]{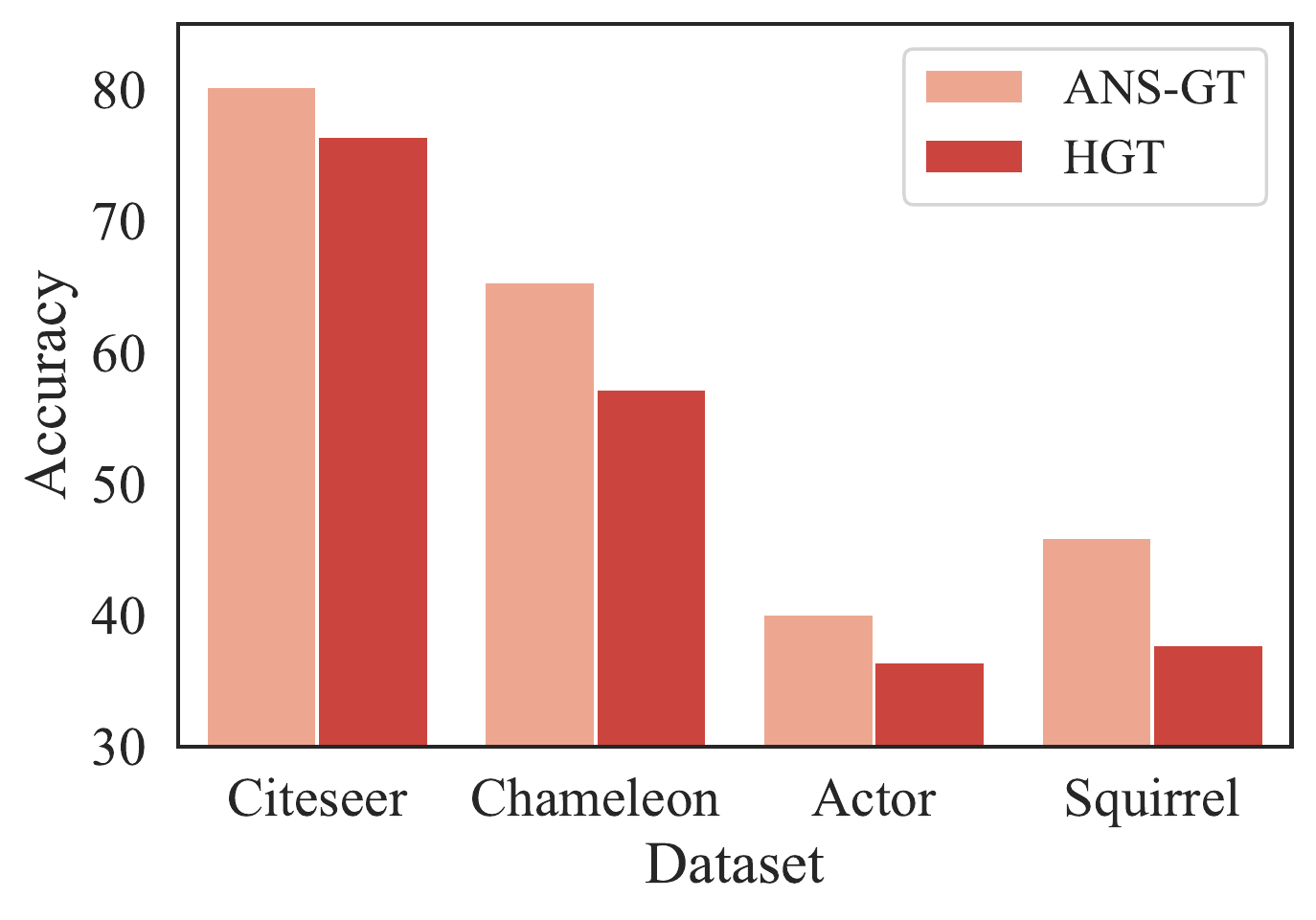}   
    \caption{Ablation studies to show the effectiveness of the adaptive node sampling module. HGT refers to the Hierarchical Graph Transformer without adaptive node sampling.}
    \label{ablation}
    \vspace{-1.0em}
\end{wrapfigure}

\textbf{Effectiveness of Adaptive Node Sampling.}
Our proposed adaptive node sampling module can adjust the weights for sampling based on the rewards as the training progresses. To evaluate its effectiveness and give more insights into the ANS module, we show the normalized sampling weights as a function of training epochs on four datasets in Figure \ref{curve}. Generally, we observe that the sampling weights of different sampling strategies are time-sensitive and gradually stabilize with the increase of the number of epochs. Interestingly, we find PPR and 1-hop neighbors achieves high weights on Cora while 2-hop neighbors dominate other sampling strategies on Squirrel. This may be explained by the fact that Cora and Squirrel are strong homophily/heterophily dataset respectively. For Citeseer and Actor, the weights of KNN firstly goes up and gradually decreases. This is probably due to the reason that nodes with similar attributes are most useful for the training at the beginning stage; local neighbors such as nodes with high PPR scores are more useful at the fine-tuning stage. Furthermore, Figure \ref{ablation} shows the ablation studies of the adaptive node sampling module. We can observe that ANS-GT has a large advantage compared to its variant without the adaptive sampling module denoted as HGT (e.g., On Chameleon, ANS-GT achieves 65.42$\%$ while HGT only has 57.20$\%$).

\begin{wraptable}{r}{6.7cm}
\centering
\scriptsize
\tabcolsep=0.1cm
\caption{Sensitivity analysis of coarsening algorithms and coarsening rate.}
\vspace{1.em}
\begin{tabular}{ccccccc}
\toprule
Dataset                 & Method                & c=0.01&c=0.05 &  \begin{tabular}[c]{@{}c@{}} c=0.10 \\ \end{tabular} & \begin{tabular}[c]{@{}c@{}} c=0.50\end{tabular}&c=1.00\\ \midrule
\multirow{3}{*}{Cora}   &VN   &\textbf{88.60}  &\bf88.55   & 88.14                  & \textbf{87.85}    &87.26             \\
                         & VE                &87.95&88.13     & \textbf{88.30}              & 87.32 &87.22            \\
                         & JC   &88.49 &88.20    & 87.46              & 87.36 &\bf87.28            \\\midrule
\multirow{3}{*}{Actor} &VN  &\bf39.72  &39.45  & \textbf{40.10}                  & 38.83     &39.08            \\
                         & VE                &39.20  &\bf39.66  & 39.51             & 38.94     &39.06        \\
                         & JC    & 39.15 &39.85   & 39.92             & \textbf{39.16}  &\bf39.09           \\
\bottomrule
\end{tabular}
\vspace{-0.1em}
\label{coarsening sensitivity}
\end{wraptable}

\textbf{Graph Coarsening Methods. }
In ANS-GT, we apply a hierarchical attention mechanism where the interactions between the center node and super-nodes generated by graph coarsening are considered. 
Here, we evaluate ANS-GT with different coarsening algorithms and different coarsening rates. Specifically, the considered coarsening algorithms include Variation Neighborhoods (VN) \cite{loukas2019graph}, Variation Edges (VE) \cite{loukas2019graph}, and Algebraic (JC) \cite{ron2011relaxation}. We vary the coarsening rate from 0.01 to 0.50. In Table \ref{coarsening sensitivity}, we can observe that there is no significant difference between different coarsening algorithms, indicating the robustness of ANS-GT w.r.t. them. As for the coarsening rate, the results indicate that the coarsening rate of 0.01 to 0.10 has the best performance.
\section{Conclusion}
Motivated by the obstacles to generalize Transformer to large graphs, we propose \textbf{A}daptive \textbf{N}ode \textbf{S}ampling for \textbf{G}raph \textbf{T}ransformer (ANS-GT), which modifies a multi-armed bandit algorithm to adaptively sample nodes for attention in this paper. To incorporate long-range dependencies and global contexts, we further design a hierarchical graph attention scheme in which coarse-grained attention is achieved with graph coarsening. We empirically evaluate our method on six benchmark datasets to show the advantage over existing Graph Transformers and popular GNNs. The detailed analysis demonstrates that the adaptive node sampling module could effectively adjust the sampling strategies according to graph properties.
Finally, We hope our work can help Transformer generalize to the graph domain and encourage the unified modeling of multi-modal data. 

\section{Acknowledgments}
This research was partially supported by grants from the National Natural Science Foundation of China (Grant No.s 61922073 and U20A20229).

\bibliographystyle{plain}
\bibliography{main}
\section*{Checklist}

\begin{enumerate}
  \item Do the main claims made in the abstract and introduction accurately reflect the paper's contributions and scope?
    \answerYes{}
  \item Did you describe the limitations of your work?
    \answerYes{Please see Appendix.}
  \item Did you discuss any potential negative social impacts of your work?
    \answerYes{Please see Appendix.}
  \item Have you read the ethics review guidelines and ensured that your paper conforms to them?
    \answerYes{}
\end{enumerate}

\begin{enumerate}
  \item Did you state the full set of assumptions of all theoretical results?
    \answerNA{}
        \item Did you include complete proofs of all theoretical results?
    \answerNA{}
\end{enumerate}

\begin{enumerate}
  \item Did you include the code, data, and instructions needed to reproduce the main experimental results (either in the supplemental material or as a URL)?
    \answerNo{The code will be released once the paper is accepted.}
  \item Did you specify all the training details (e.g., data splits, hyperparameters, how they were chosen)?
    \answerYes{}
        \item Did you report error bars (e.g., with respect to the random seed after running experiments multiple times)?
    \answerYes{}
        \item Did you include the total amount of computing and the type of resources used (e.g., type of GPUs, internal cluster, or cloud provider)?
    \answerYes{}
\end{enumerate}

\begin{enumerate}
  \item If your work uses existing assets, did you cite the creators?
    \answerYes{}
  \item Did you mention the license of the assets?
    \answerNA{}
  \item Did you include any new assets either in the supplemental material or as a URL?
    \answerNA{}
  \item Did you discuss whether and how consent was obtained from people whose data you're using/curating?
    \answerNA{}
  \item Did you discuss whether the data you are using/curating contains personally identifiable information or offensive content?
    \answerNA{}
\end{enumerate}

\begin{enumerate}
  \item Did you include the full text of instructions given to participants and screenshots, if applicable?
    \answerNA{}
  \item Did you describe any potential participant risks, with links to Institutional Review Board (IRB) approvals, if applicable?
    \answerNA{}
  \item Did you include the estimated hourly wage paid to participants and the total amount spent on participant compensation?
    \answerNA{}
\end{enumerate}


\appendix

\newpage

\section{More Details of Motivating Observations}
\label{details of observation}
\begin{figure}[ht]
	\centering
	\subfigure[$\alpha = 0.75$]{\includegraphics[width=0.32\linewidth]{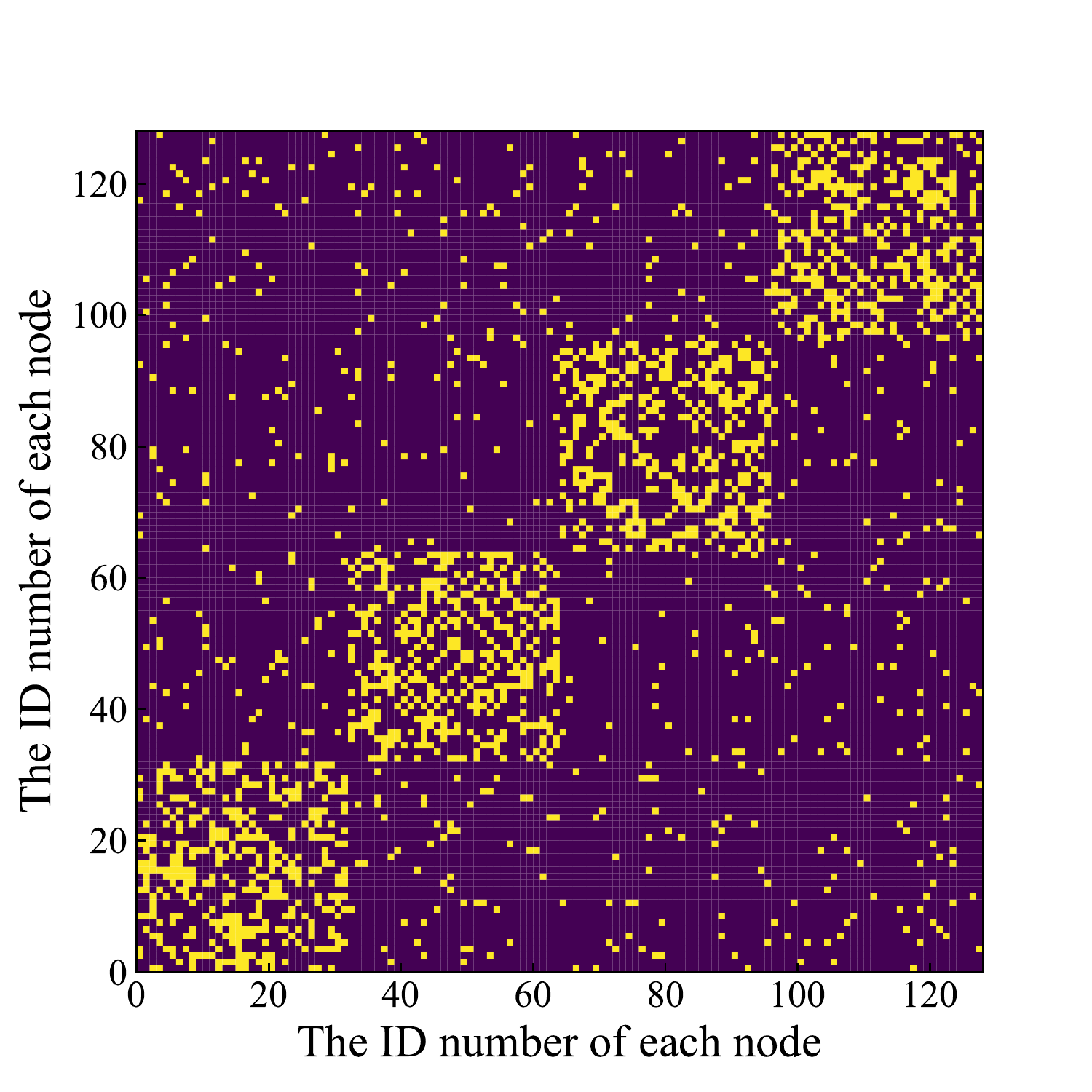}}
    \subfigure[$\alpha = 0.25$]{\includegraphics[width=0.32\linewidth]{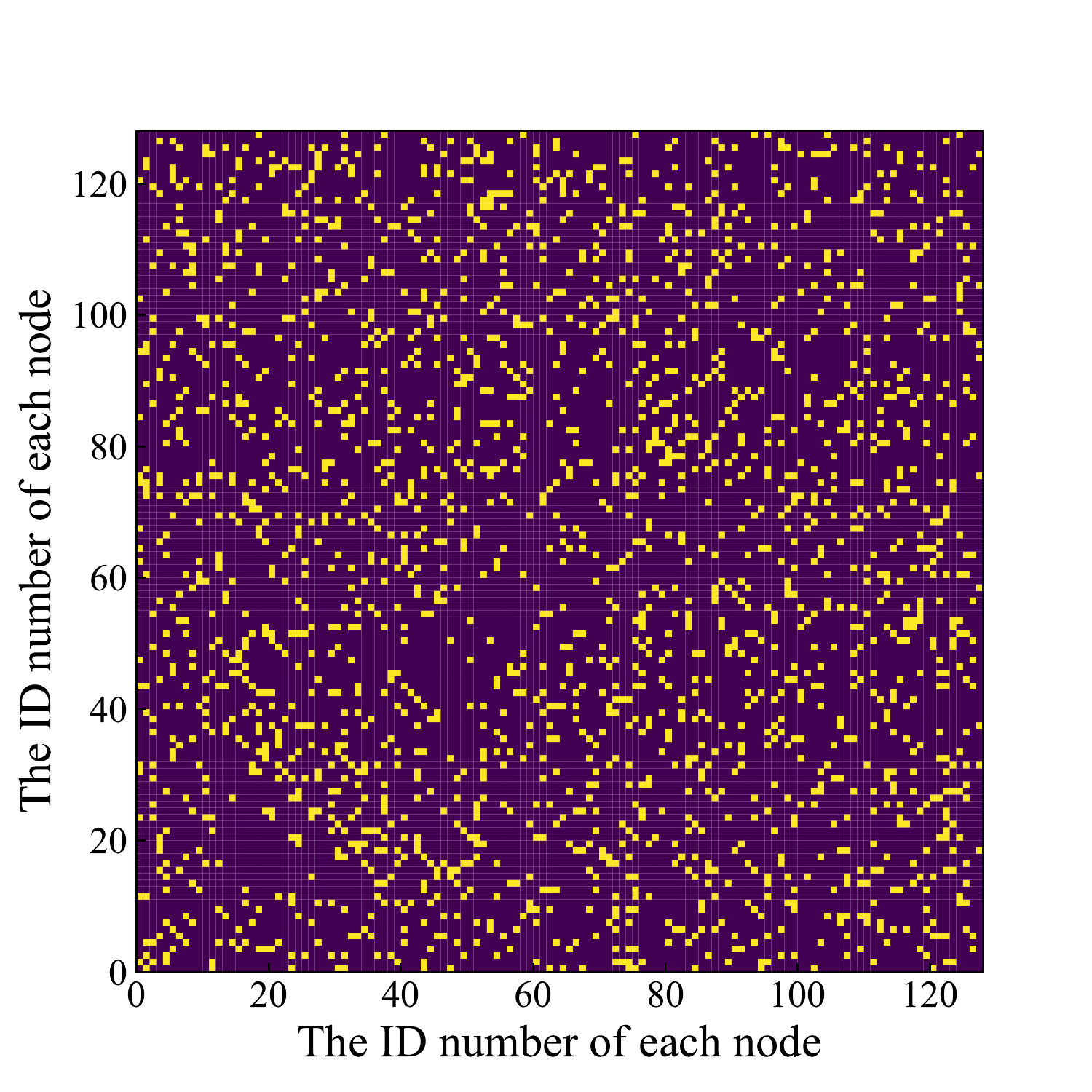}}
    \subfigure[$\alpha = 0.05$]{\includegraphics[width=0.32\linewidth]{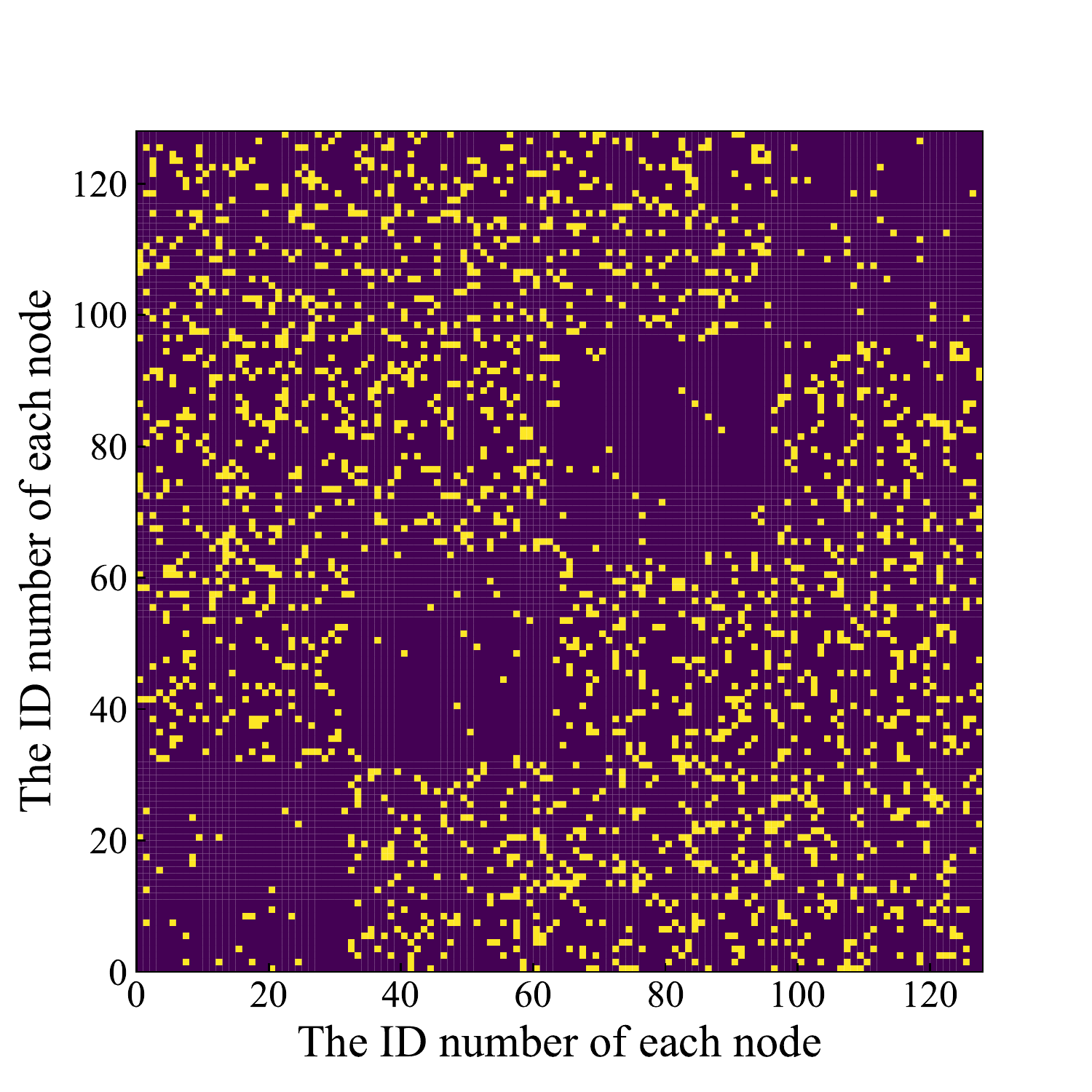}}
	\caption{The adjacency matrix of the Newman network with strong homophily, randomness and strong heterophily respectively. (The yellow dots indicate connected edges and the purple dots indicate no edges.)}
	\label{adj}
\end{figure}

\textbf{Experiment Setup.} We conduct experiments on the Newman artificial networks \cite{girvan2002community} with different properties. The network consists of 128 nodes divided into 4 classes, where each node has on average $z_{in}$ edges (i.e.,
intra-class edges) connecting to nodes of the same
class and $z_{out}$ edges (i.e., inter-class edges) to nodes of other classes, and $z_{in} + z_{out} = 16$. Here two indicators are used: $\rho_{in} = z_{in}/32$ and $\rho_{out} = z_{out}/96$, to indicate the graph property, i.e., $\rho_{in} > \rho_{out}, \rho_{in} = \rho_{out}$ and $\rho_{in} <
\rho_{out}$ means the graph with homophily, randomness and heterophily, respectively. In Figure \ref{adj}, we show the visualization of the adjacency matrix with strong homophily, randomness and strong heterophily.

For the node attributes, we generate $4h$-dimensional binary attributes (i.e., $x_i$) for each node to form 4
attribute clusters, corresponding to the 4 classes \cite{he2017joint}.
To be specific, for every node in the $i$-th class, we use a binomial distribution with mean $p_{in} = h_{in}/h$ to generate a $h$-dimensional binary vector as its $((i-1)\times h + 1)$-th to $(i\times h)$-th attributes, and generated the rest attributes using a binomial distribution with mean $p_{out} = h_{out}/(3h)$. In our
experiments, we set $4h$ = 200 and $h_{out} = 4 (h_{in} +h_{out} = 16)$, so that $p_{in} > p_{out}$, the $h$-dimensional attributes are associated with the $i$-th class with a higher probability, whereas the rest $3h$ attributes are irrelevant. For the model implementation, we use the Gophormer \cite{zhao2021gophormer} with the default setting for the demonstration. For each center node, we sample 10 nodes with 1-hop, 2-hop, KNN and PPR strategies 16 times for data augmentation.

\section{Dataset Statistics}
In Table \ref{tab: datasets}, we show the detailed statistics of 9 datasets. 
\begin{table}[ht]
\centering
\caption{The statistics of the datasets.}
\begin{tabular}{c|cccccc} \toprule
Dataset     & \#Nodes & \#Edges & \#Classes & \#Features & Type & $\alpha$\\ \midrule
Cora        & 2,708    & 5,278    & 7         & 1,433       & Citation network   & 0.83\\
Citeseer    & 3,327    & 4,522    & 6         & 3,703       & Citation network   & 0.71 \\
Pubmed      & 19,717   & 44,324   & 3         & 500        & Citation network   & 0.79 \\
Chameleon        & 2,277   & 31,371   & 5         & 2,325       & Wiki pages &  0.23\\
Actor & 7,600    & 26,659  & 5         & 932  & Actors in movies  &  0.22\\
Squirrel      & 5,201    & 198,353  & 5  & 2,089      & Wiki pages & 0.22
\\ 
Texas &183 &279 & 5&1703&Web pages&0.11\\
Cornell &183 &277 & 5&1703&Web pages&0.30\\
Wisconsin &251 &499 & 5&1703&Web pages&0.21\\
\bottomrule
\end{tabular}

\label{tab: datasets}
\end{table}

\section{Additional Results and Analysis}
\label{additional analysis}
\begin{figure}[t]
	\centering
	\subfigure[]{\includegraphics[width=0.24\linewidth]{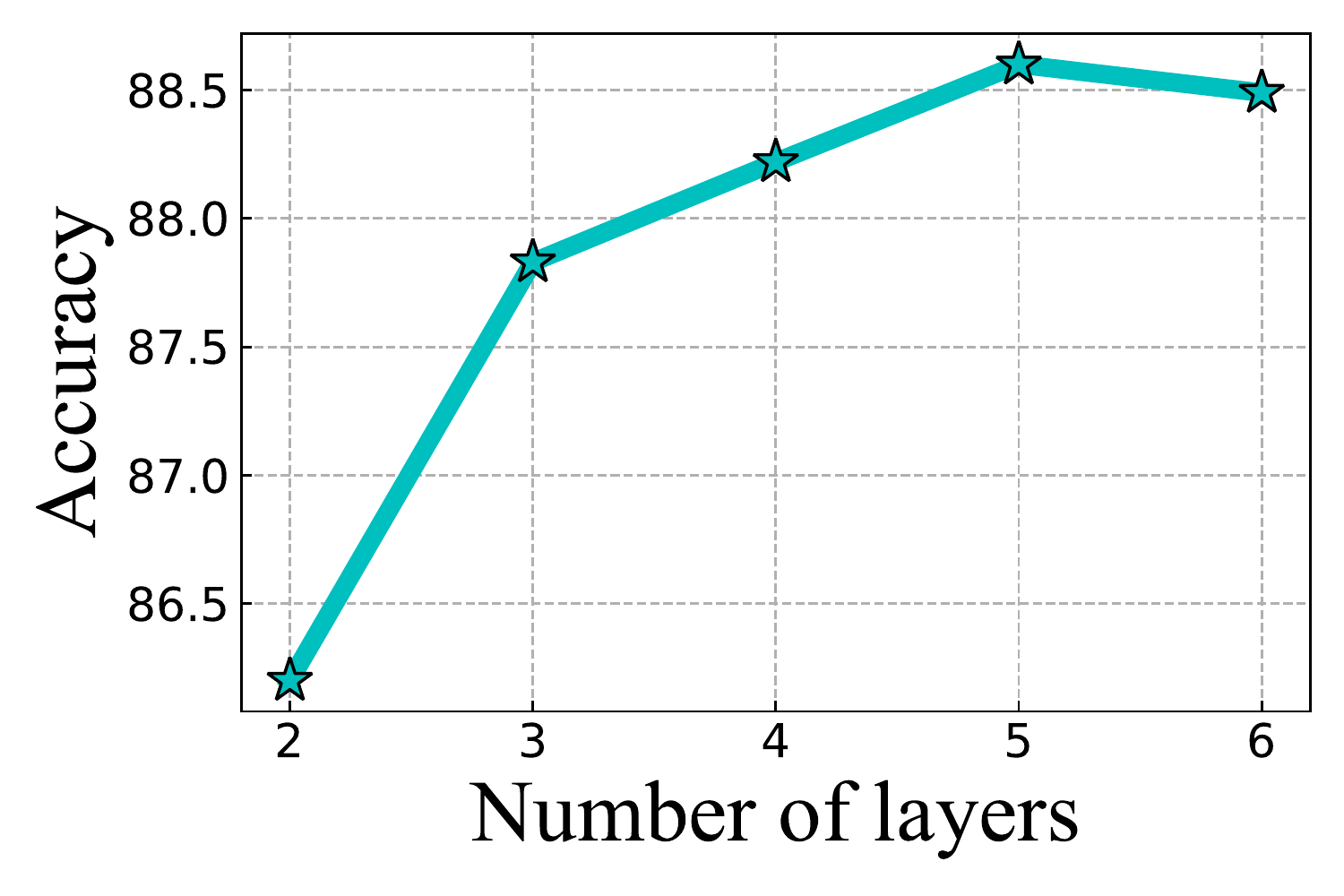}}
    \subfigure[]{\includegraphics[width=0.24\linewidth]{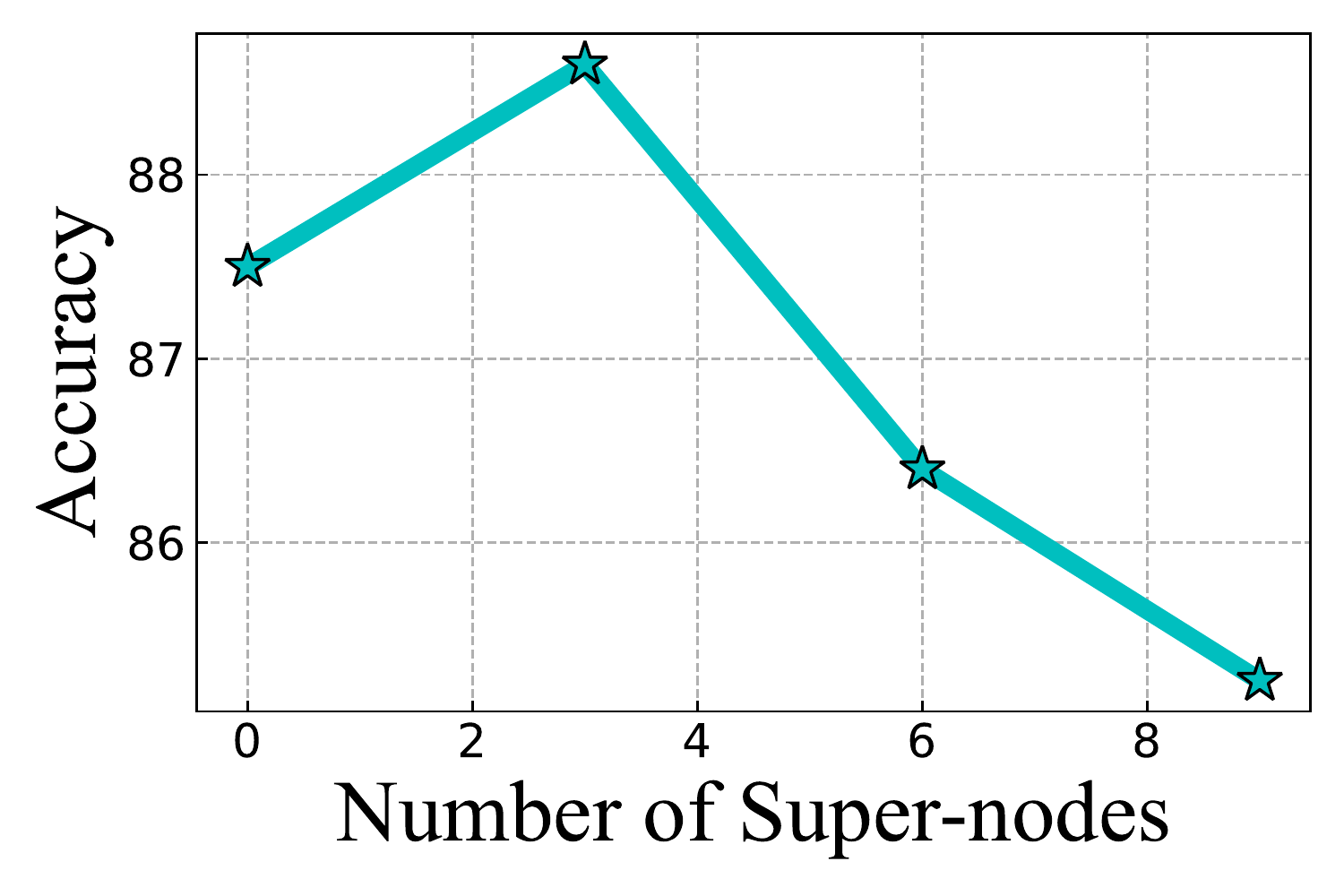}}
    \subfigure[]{\includegraphics[width=0.24\linewidth]{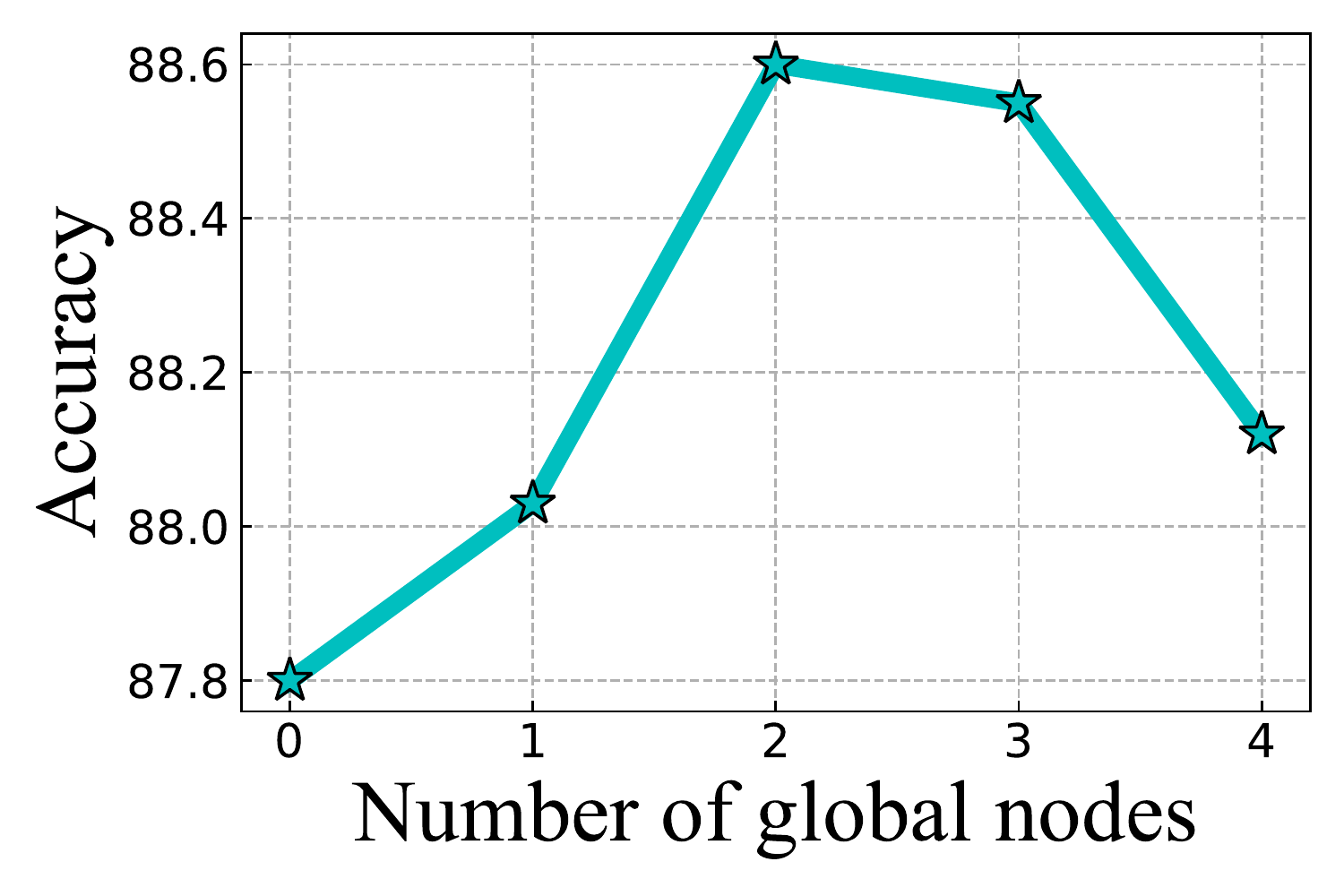}}
    \subfigure[]{\includegraphics[width=0.24\linewidth]{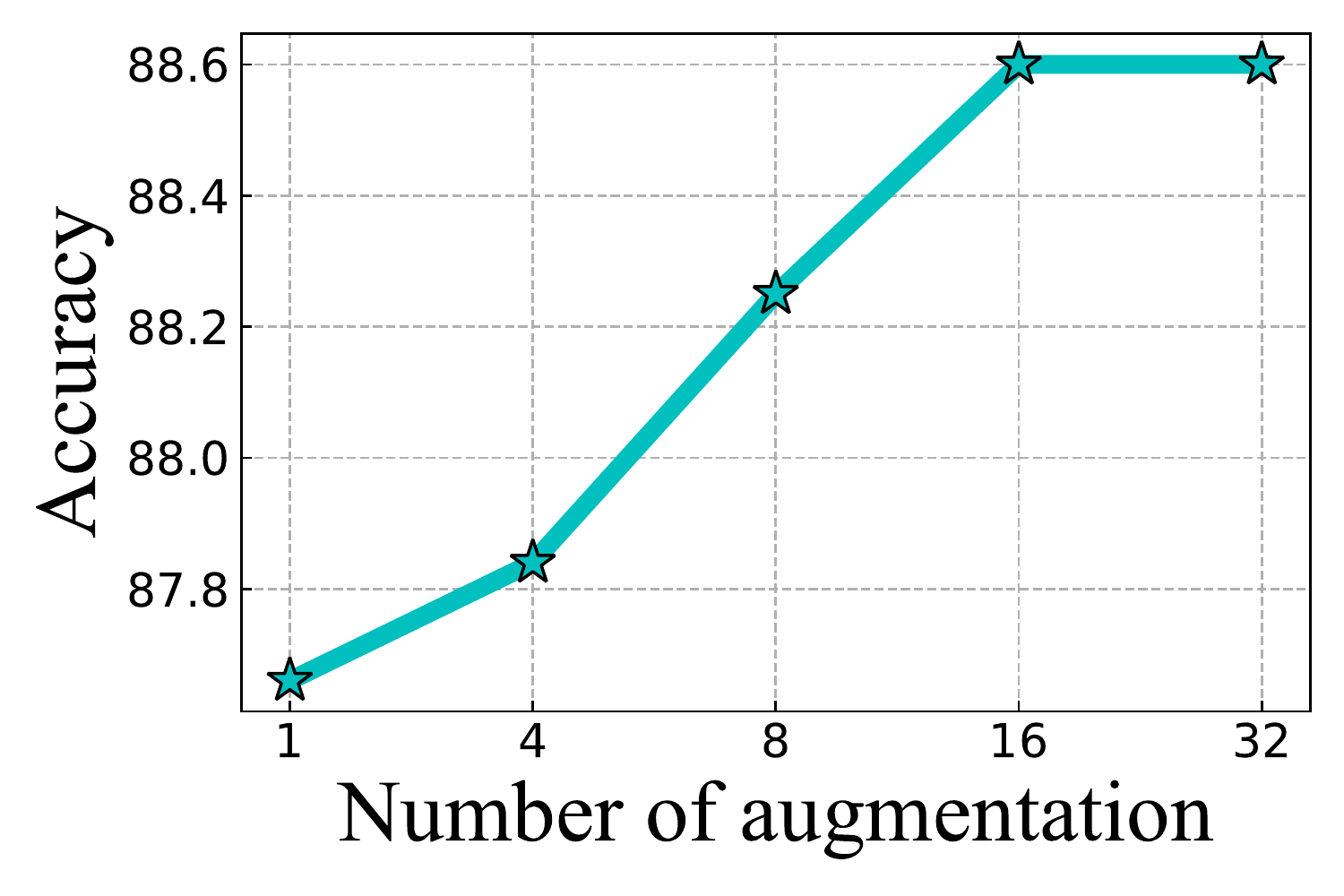}}
	\caption{Parameter sensitivity analysis on Cora. We show (a) the influence of the number of layers; (b) the number of super-nodes; (c) the number of global nodes; (d) and the number of augmentation.}
	\label{parameter analysis}
\end{figure}

\begin{figure}[t]
	\centering
	\subfigure[]{\includegraphics[width=0.24\linewidth]{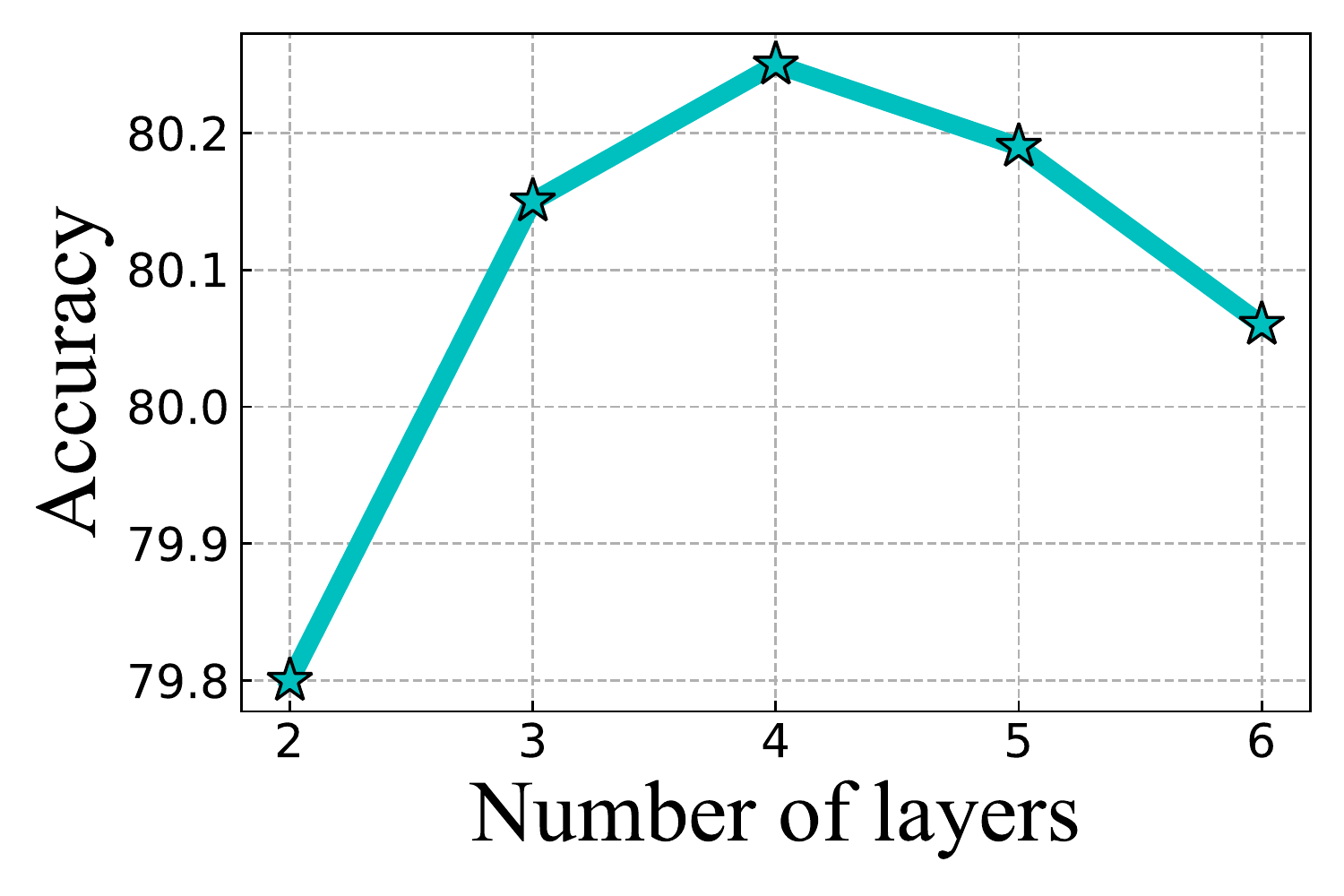}}
    \subfigure[]{\includegraphics[width=0.24\linewidth]{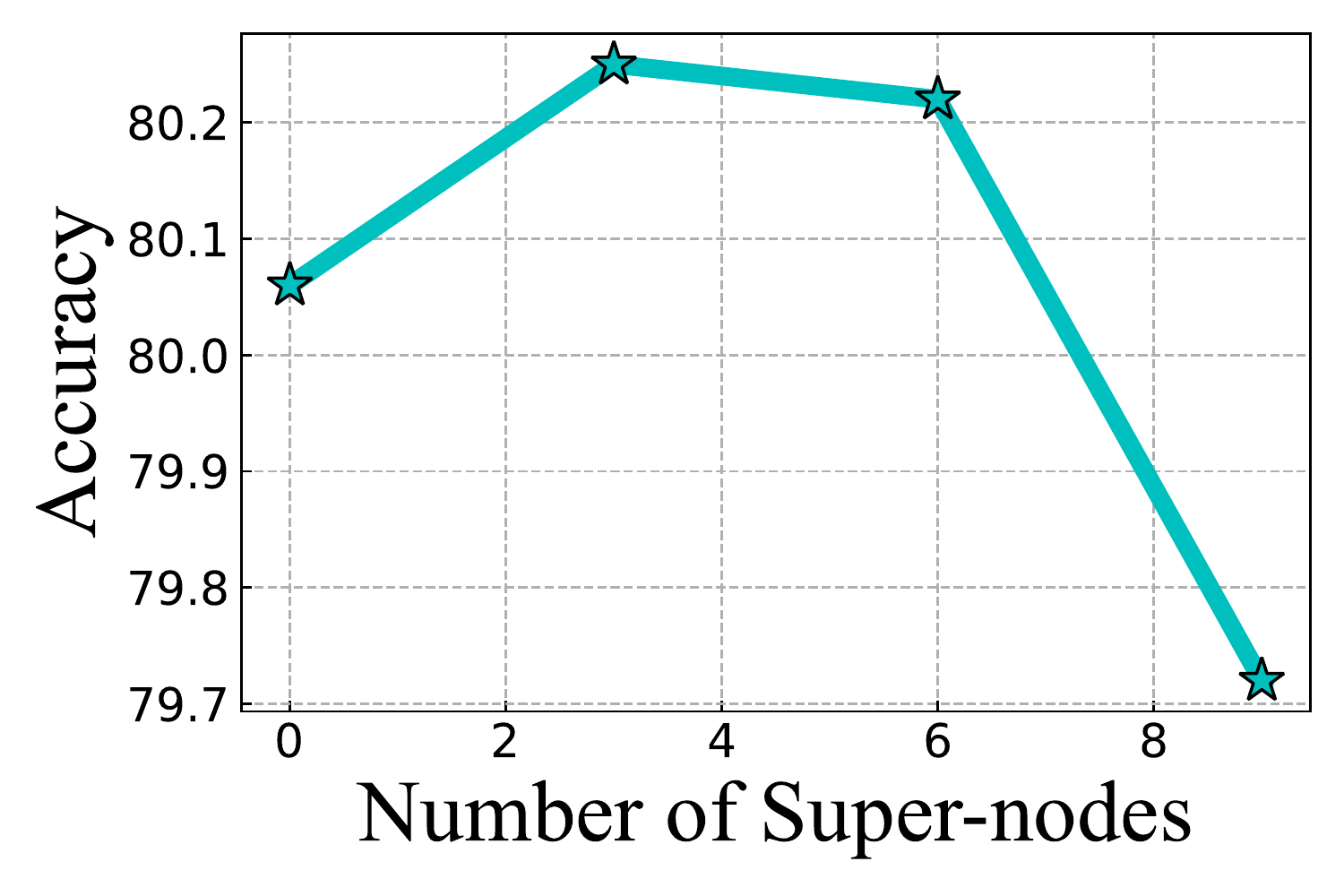}}
    \subfigure[]{\includegraphics[width=0.24\linewidth]{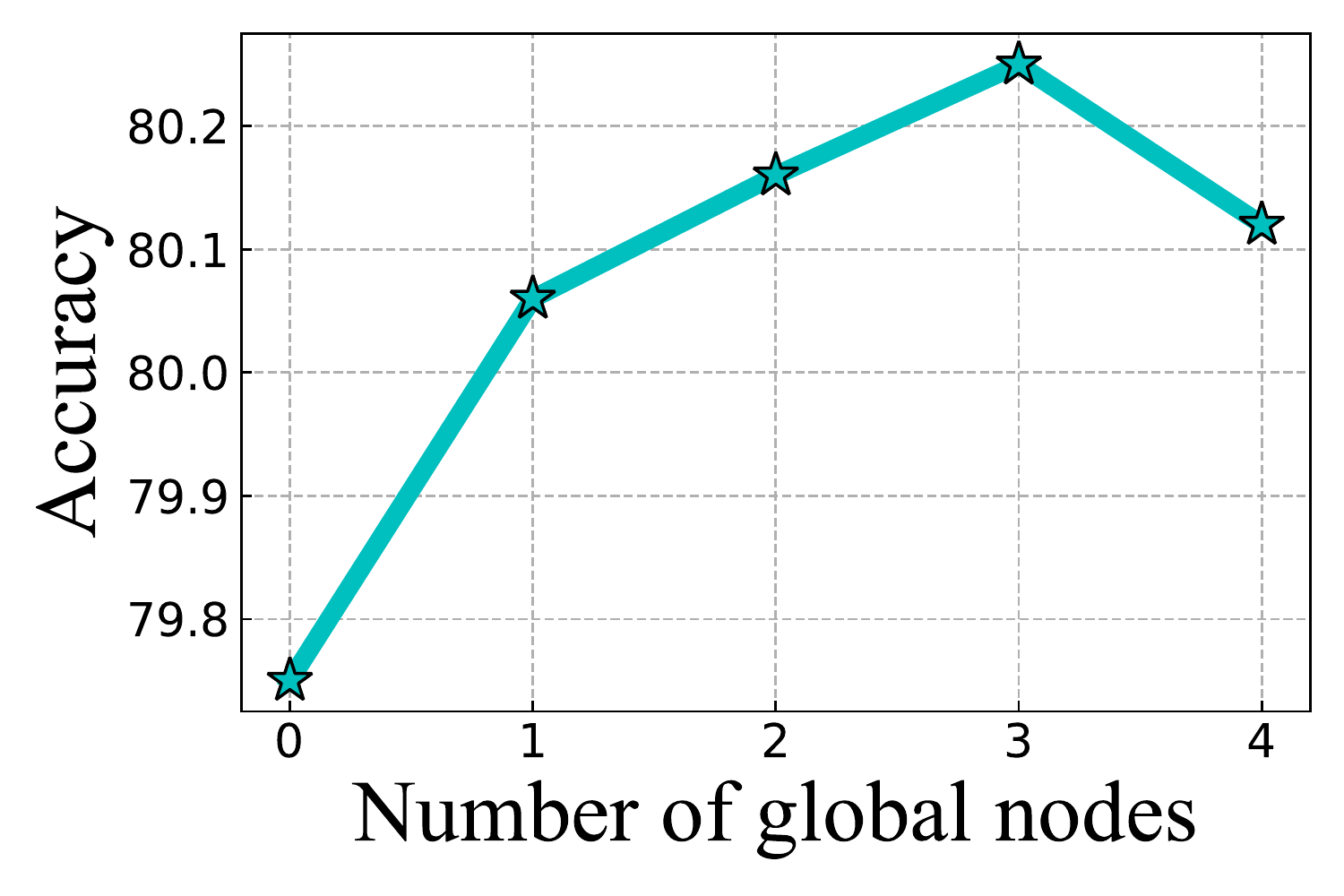}}
    \subfigure[]{\includegraphics[width=0.24\linewidth]{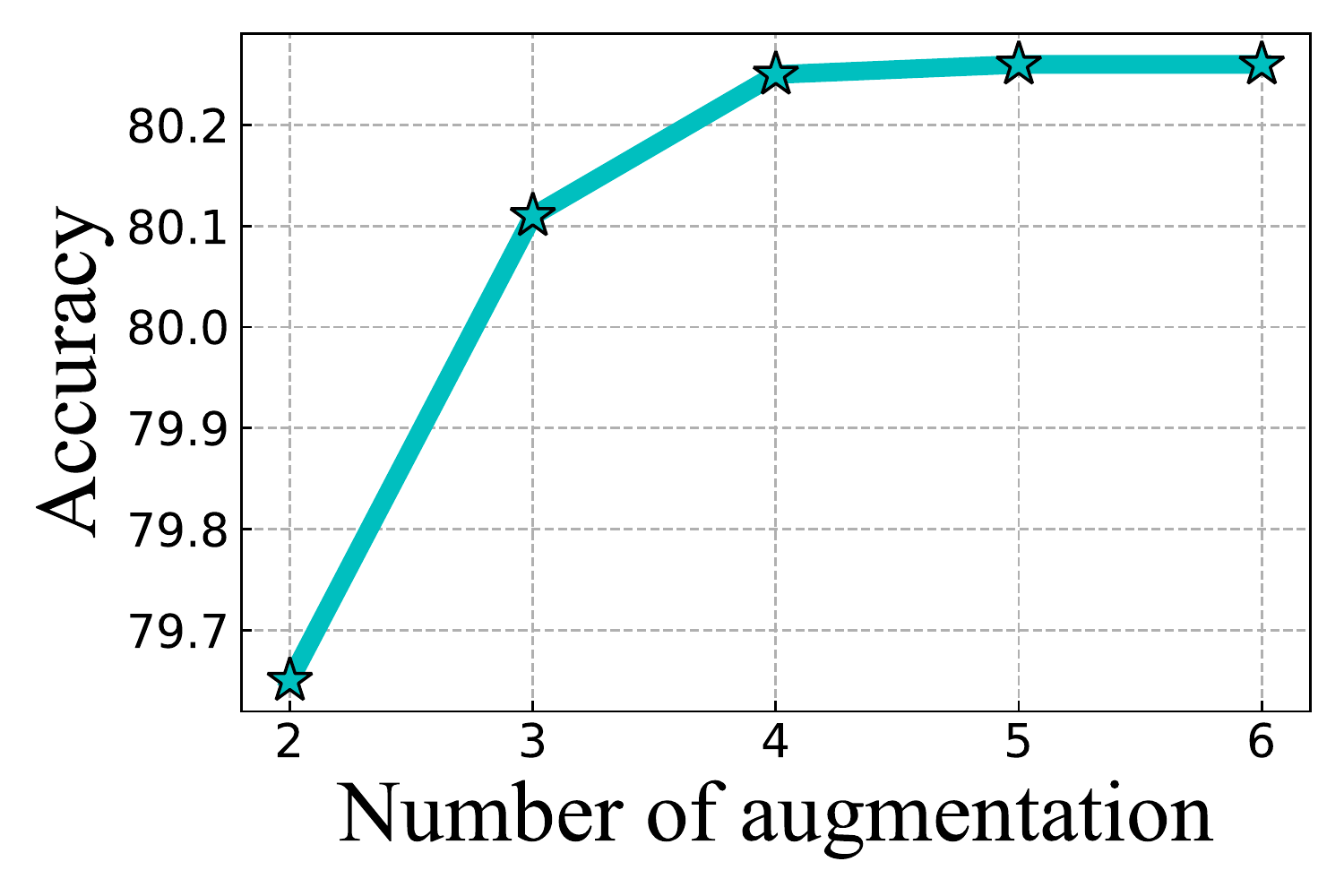}}
	\caption{Parameter sensitivity analysis on Citeseer.}
\end{figure}

\begin{figure}[t]
	\centering
	\subfigure[]{\includegraphics[width=0.24\linewidth]{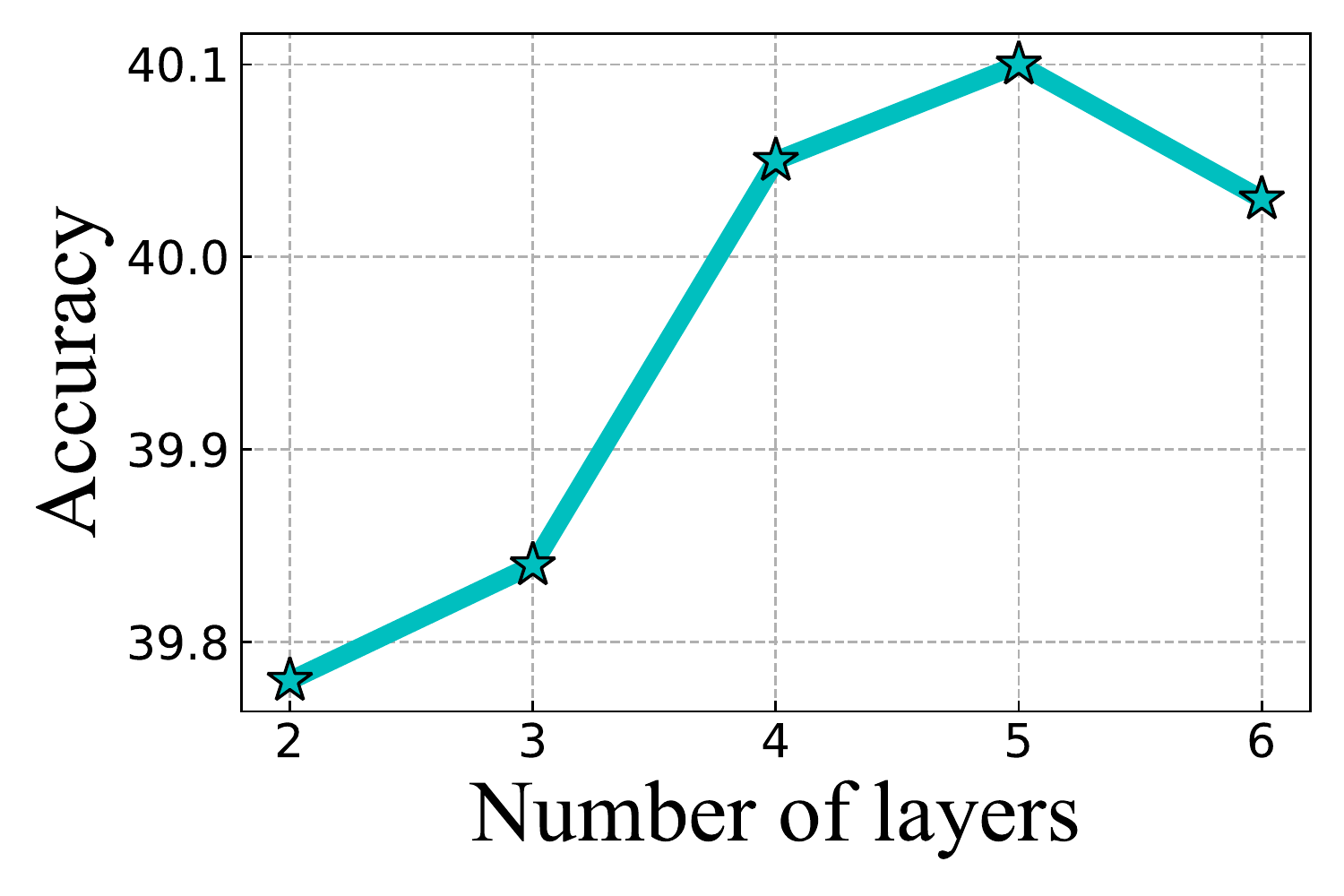}}
    \subfigure[]{\includegraphics[width=0.24\linewidth]{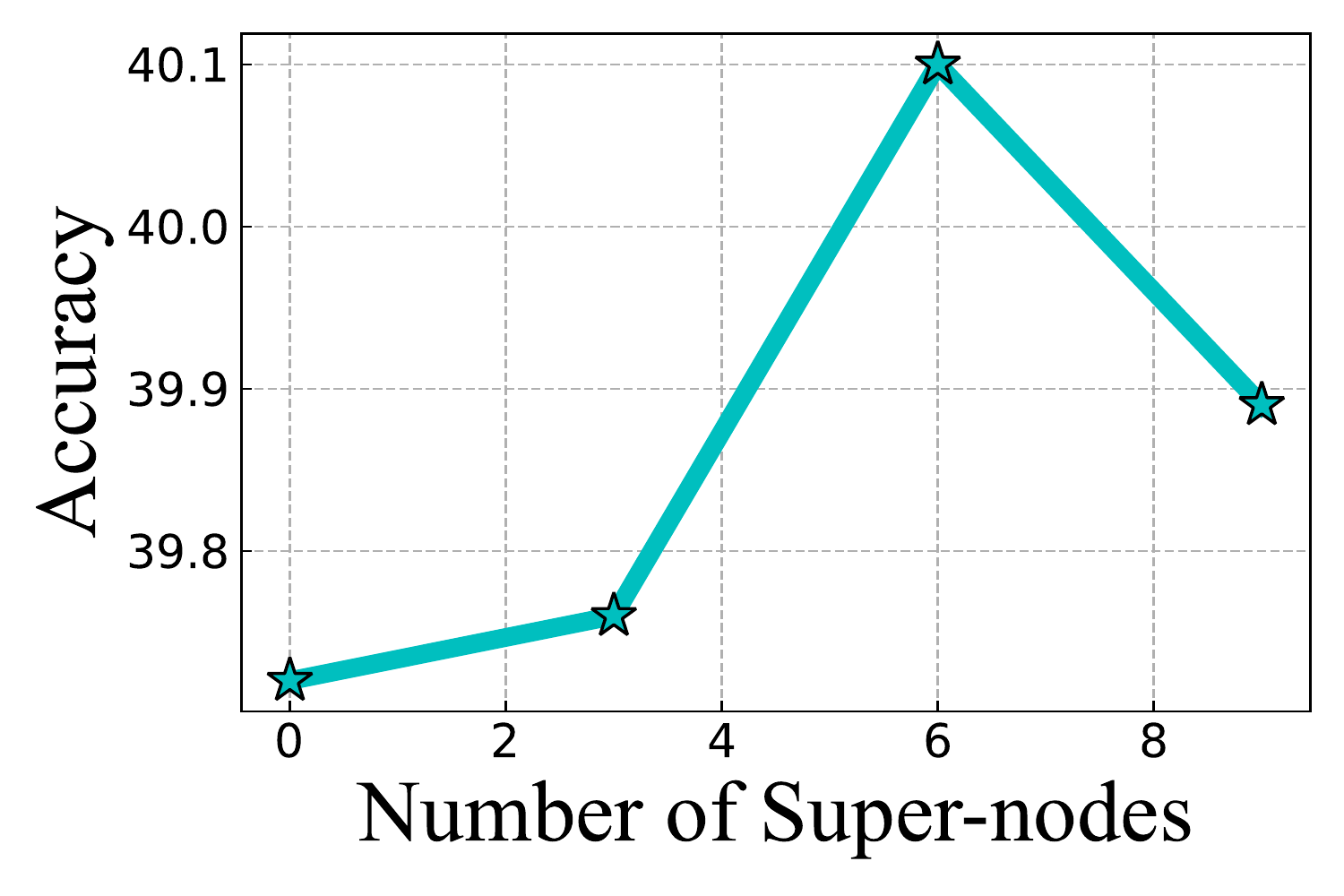}}
    \subfigure[]{\includegraphics[width=0.24\linewidth]{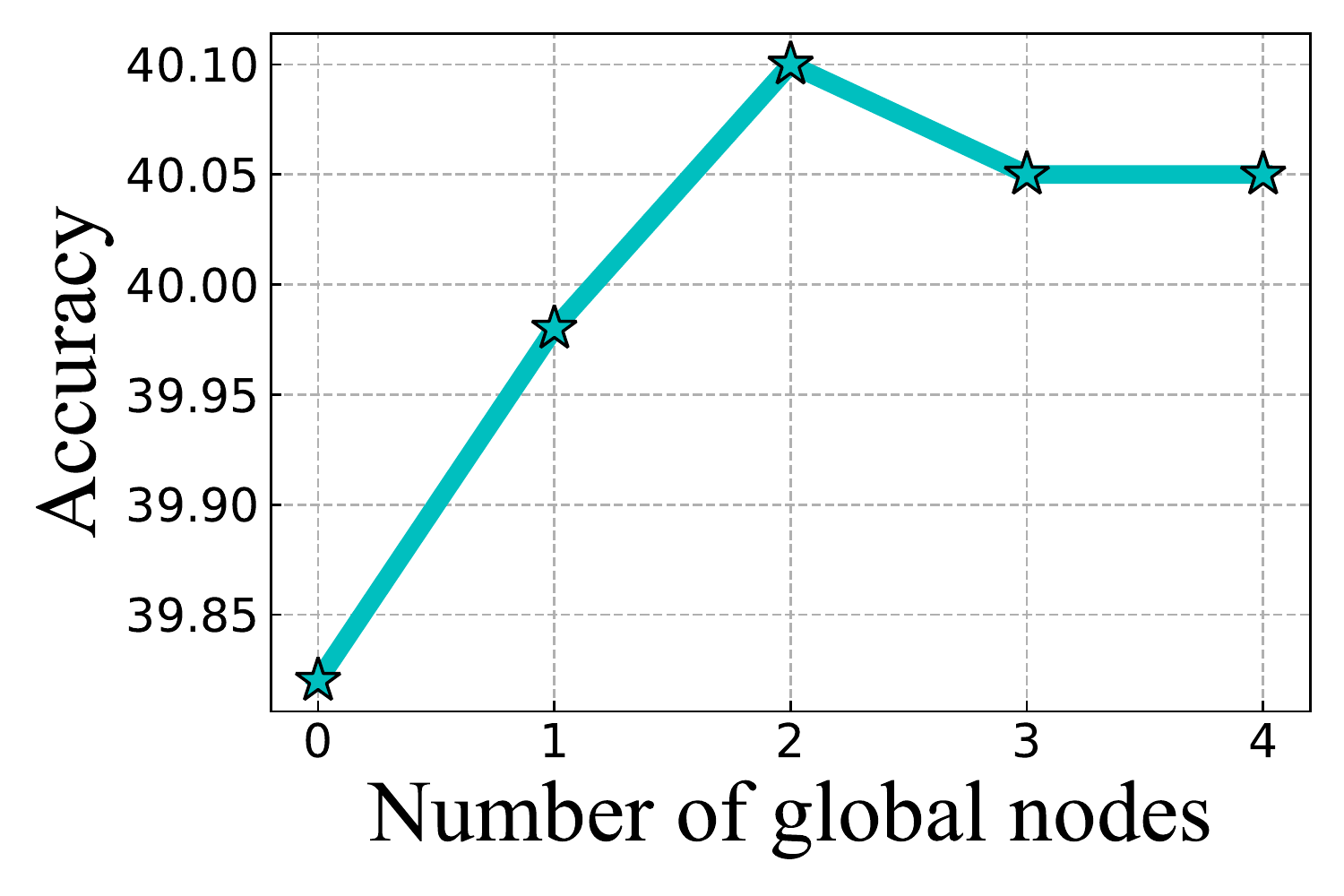}}
    \subfigure[]{\includegraphics[width=0.24\linewidth]{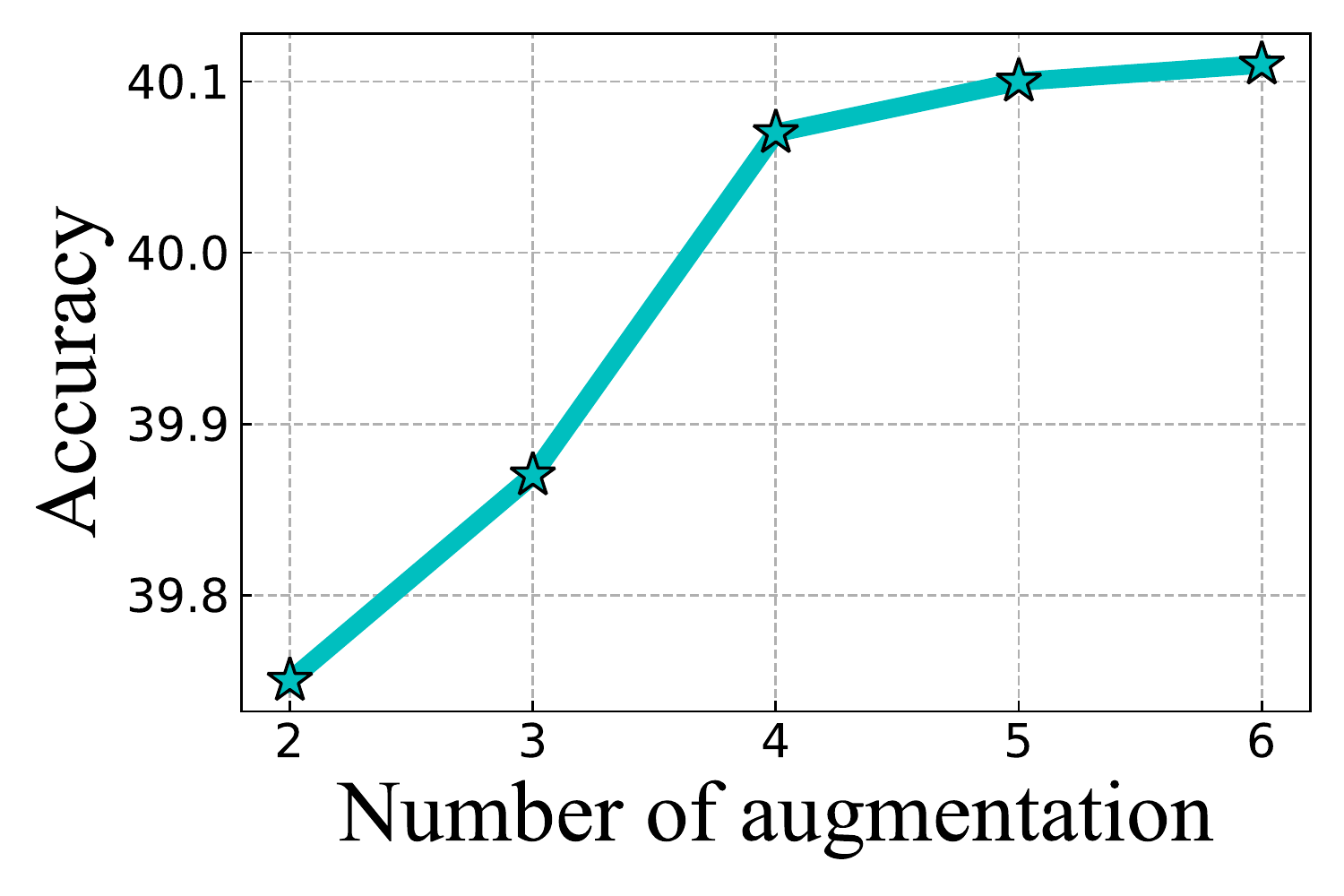}}
	\caption{Parameter sensitivity analysis on Actor. }
\end{figure}

\begin{table}[ht]
\centering
\caption{Time consumption for graph coarsening (s). The coarsening rate $c$ is defined as $\frac{|V'|}{|V|}$}
\begin{tabular}{ccccc}
\toprule
Dataset                 & Method                & c=0.01 &  \begin{tabular}[c]{@{}c@{}} c=0.10 \\ \end{tabular} & \begin{tabular}[c]{@{}c@{}} c=0.50\end{tabular}\\ \midrule
\multirow{3}{*}{Cora}   &VN   &3.792    & 3.536                  & 2.215                 \\
                         & VE                &1.540     & 1.516              & 0.851             \\
                         & JC   &1.454     & 1.271              & 0.665             \\\midrule
\multirow{3}{*}{Actor} &VN  &11.868    &  11.53                 & 7.000                 \\
                         & VE                &7.535    & 6.911             & 3.154             \\
                         & JC    &11.785    & 11.624             & 3.651             \\
\bottomrule
\end{tabular}
\label{coarsening time}
\end{table}
\begin{table}[t]
\centering
\caption{Time consumption for adaptive node sampling per epoch (s).}
\begin{tabular}{c|cccccc} \toprule
Dataset     & Cora & Citeseer & Pubmed & Chameleon & Actor & Squirrel\\ \midrule
Time         & 0.838    & 0.926    & 1.717  & 0.855       & 1.026   & 0.977\\
 \bottomrule
\end{tabular}
\label{tab: ANS time}
\end{table}

\begin{table}[t]
\centering
\caption{Efficiency comparisons with Graph Transfomer baselines. The average training time per epoch (s)}
\begin{tabular}{c|cccccc} \toprule
Dataset     & Cora & Citeseer & Pubmed & Chameleon & Actor & Squirrel\\ \midrule
Graphormer        & 25.670   & 37.899   & 26.436   & 26.343 & 30.105   & 29.771\\
Gophormer   & 11.210   & 12.121   & 16.116   & 10.305 & 12.243   & 12.579 \\
ANS-GT     & 11.495   & 12.143   & 16.270   & 10.311 & 12.240   & 12.571 \\
\bottomrule
\end{tabular}
\label{training time}
\end{table}
\begin{table}[t]
\centering
\caption{The performance of ANS-GT and selected baselines on OGB datasets}
\begin{tabular}{c|cccc} \toprule
Methods     & GCN &	GraphSAGE &	GPRGNN  &ANS-GT\\ \midrule
ogbn-arxiv        & 71.72±0.45	&71.46±0.26&	70.90±0.23&	\bf72.84±0.34\\
ogbn-products   & 75.57±0.28 &	78.61±0.31 &	79.76±0.59&	\bf82.15±0.30 \\
\bottomrule
\end{tabular}
\label{ogb}
\end{table}

\subsection{Hyper-parameter Analysis}

In Figure \ref{parameter analysis}, we study the sensitivity of ANS-GT on four important hyper-parameters: the number of transformer layers, the number of super-nodes $n_s$, the number of global nodes $n_g$, and the number of data augmentation $\mathcal{S}$. In Figure \ref{parameter analysis} (a), we observe that the performance increases at the beginning with the increase of transformer layers. The reason is that stacking more transformer layers improves the model's capability. However, we witness a slight performance decrease when the number of layers exceeds 6, possibly suffering from over-fitting. Figure \ref{parameter analysis} (b) and (c) presents the node classification performance with $n_s$ varying from 0 to 9 and $n_g$ from 0 to 4 respectively. With the increase of $n_s$ and $n_g$, the performance increases until reaches a peak and then decreases. This is expected as the optimal number of super-nodes and global nodes help incorporate long-range dependencies and global context in the graph while too large $n_s$ and $n_g$ lead to redundant noise. Hence, the number of super-nodes and global nodes should be carefully chosen to achieve optimal performance. Finally, we show the influence of the number of data augmentation in Figure \ref{parameter analysis} (d). With the increase of $\mathcal{S}$, the node classification performance improves steadily until stabilizes. The results indicate data augmentation in the training and the bagging aggregation in the inference can effectively improve the classification accuracy. In conclusion, we recommend 5 transformer layers, 3 super-nodes, 2 global nodes, and an augmentation number of 16 for Cora.

\subsection{Efficiency Analysis of ANS-GT}
Here we show more experiment results and analysis on the efficiency of ANS-GT.

In Table \ref{coarsening time}, we present the time consumption of executing the graph coarsening algorithm on Cora and Actor with different coarsening rates and methods. Since graph coarsening only needs to be done once at the pre-processing stage, the time consumption is acceptable.

In Table \ref{tab: ANS time}, we show the time consumption for adaptive sampling in one epoch. In our algorithm, we update the sampling weights every $T$ epochs ($T = 100$ in experiments). Hence, the time cost of the adaptive node sampling module is trivial. 

In Table \ref{training time}, we present the training efficiency comparisons with other Graph Transformer baselines. Specifically, we show the average training time per epoch. As can be observed in Table \ref{training time}, ANS-GT has comparable training time with Gophormer and its efficiency is much better than Graphormer. 

\subsection{Limitations and Potential Negative Social Impacts}
One limitation of our work is that it introduces more hyper-parameters for finetuning. Since our work utilizes adaptive node sampling, it may lead to potential biases in sampling nodes for training.
\subsection{Additional Results on OGB Datasets}
We additionally try ANS-GT on ogbn-arxiv and ogbn-products datasets from OGB \cite{hu2020ogb}, which contains 169,343 and 2,449,029 nodes respectively. We use the official train/valid/test split and data pre-processing details can be found in \cite{hu2020ogb}. The model setup of ANS-GT follows Section 6.1. Three competitive baselines including GCN, GraphSAGE, and GPRGNN are selected. We present average accuracies and standard deviations over 5 runs in Table \ref{ogb}. Our results overperform the baselines and demonstrate the effectiveness of ANS-GT on large-scale graphs.

\section{Supplementary Information of Graph Coarsening}
 In this paper, we use 3 popular graph coarsening algorithms: Variation Neighborhoods (VN) \cite{loukas2019graph}, Variation Edges (VE) \cite{loukas2019graph}, and Algebraic JC (JC) \cite{ron2011relaxation}. VN and VE belong to the local variation algorithms which coarsen graphs by preserving the spectral properties of adjacency matrix. Local variation algorithms differ only in the type of contraction sets that they consider: Variation Edges only contracts edges, whereas contraction sets in Variation Neighborhoods are subsets of nodes’ neighborhood. In Algebraic JC, the algebraic distances between neighboring nodes are calculated and close nodes are contracted to form clusters. More information of the coarsening algorithms can be found in their original papers.

\section{Further Discussions of ANS-GT}
In ANS-GT, we formulate the optimization strategy of node sampling in Graph Transformer as an adversary bandit problem. Specifically, ANS-GT optimizes the weights of chosen sampling heuristics instead of directly predicting the adjacent nodes to attend. Then, ANS-GT combines the weighted sampling heuristics to sample informative nodes. We did not incorporate hierarchical attention as part of the bandit learning because it samples supernodes from the coarsened graph instead of sampling nodes like the 4 strategies (1-/2- hops, KNN, and PPR).
We do not directly predict nodes to attend (e.g., using linear layers to predict informative nodes).
Directly predicting informative nodes for attention requires too much computational overhead and is hard to optimize. Comparatively, the pre-defined node sampling heuristics in our strategy help narrow the search space with prior knowledge. Moreover, the sampling strategy in ANS-GT can generalize to all nodes in the graph efficiently. Experiment results show that our strategy is effective and efficient.
\end{document}